\documentclass[letterpaper]{article} 
\usepackage{aaai2026}  
\usepackage{times}  
\usepackage{helvet}  
\usepackage{courier}  
\usepackage[hyphens]{url}  
\usepackage{graphicx} 
\usepackage{natbib}  
\usepackage{caption} 
\usepackage{algorithm}
\usepackage{algorithmic}

\usepackage{newfloat}
\usepackage{listings}
\DeclareCaptionStyle{ruled}{labelfont=normalfont,labelsep=colon,strut=off} 
\floatstyle{ruled}
\newfloat{listing}{tb}{lst}{}
\floatname{listing}{Listing}
\usepackage{booktabs}
\usepackage{epsfig}
\usepackage{physics}
\usepackage{caption}
\usepackage{subcaption}
\usepackage{multirow}
\usepackage{array}
\usepackage{enumerate}
\usepackage{algorithm}
\usepackage{algorithmic}
\usepackage{makecell}

\usepackage{xspace}
\usepackage{amssymb}
\usepackage{siunitx}
\usepackage{siunitx}
\AtBeginDocument{\RenewCommandCopy\qty\SI}

\DeclareMathOperator*{\argmin}{arg\,min}

\makeatletter
\DeclareRobustCommand\onedot{\futurelet\@let@token\@onedot}
\def\@onedot{\ifx\@let@token.\else.\null\fi\xspace}

\def\eg{\emph{e.g}\onedot} 

\def\ie{\emph{i.e}\onedot} 

\def\cf{\emph{cf}\onedot}

\def\etal{\emph{et al}\onedot}

\AtEndPreamble{
    \usepackage[capitalize]{cleveref}
    \crefname{section}{Sec.}{Secs.}
    \crefname{table}{Tab.}{Tabs.}
    \crefname{chapter}{Chapter}{Chapters}
    \crefname{figure}{Fig.}{Figs.}
    \crefname{algorithm}{Alg.}{Algs.}
}

\newcommand{\emphalg}[1]{{\color[rgb]{0.0,0.2,0.8}\textit{#1}}}
\newcommand{\commentalg}[1]{\hfill{\color[rgb]{0.0,0.6,0.3}\textit{$\triangleright$ #1}}}

\title{Robust Long-term Test-Time Adaptation for 3D Human Pose Estimation through Motion Discretization}
\author{
    Yilin Wen, Kechuan Dong, Yusuke Sugano
}
\affiliations{
    The University of Tokyo\\
    \{fylwen, kchdong, sugano\}@iis.u-tokyo.ac.jp
}

\begin{document}

\maketitle

\begin{abstract}

Online test-time adaptation addresses the train-test domain gap by adapting the model on unlabeled streaming test inputs before making the final prediction.
However, online adaptation for 3D human pose estimation suffers from error accumulation when relying on self-supervision with imperfect predictions, leading to degraded performance over time.  
To mitigate this fundamental challenge, we propose a novel solution that highlights the use of motion discretization.
Specifically, we employ unsupervised clustering in the latent motion representation space to derive a set of anchor motions, whose regularity aids in supervising the human pose estimator and enables efficient self-replay.
Additionally, we introduce an effective and efficient soft-reset mechanism by reverting the pose estimator to its exponential moving average during continuous adaptation.
We examine long-term online adaptation by continuously adapting to out-of-domain streaming test videos of the same individual, which allows for the capture of consistent personal shape and motion traits throughout the streaming observation. 
By mitigating error accumulation, our solution enables robust exploitation of these personal traits for enhanced accuracy.
Experiments demonstrate that our solution outperforms previous online test-time adaptation methods and validate our design choices.

\end{abstract}

\section{Introduction}

Estimating 3D human body pose is essential for interpreting human behaviors.
Given streaming video input, accurate pose estimation enables prompt evaluation of user performance, facilitating applications such as human-robot collaboration, physical training, and immersive interactions.
Despite recent advances of learning-based methods ~\cite{kolotouros2019learning,kocabas2020vibe,goel2023humans}, pre-trained pose estimators often suffer from performance degradation in real-world scenarios that deviate from their training domains.

To address this issue, recent studies have explored online test-time adaptation. 
During test time, given the input unlabeled streaming video, these methods employ self-supervised learning to continuously update the model.
This allows for online 3D pose estimation for each incoming batch while gradually adapting to the test domain, aiming to capture domain-specific patterns for enhanced accuracy over time.
For example, BOA~\cite{guan2021bilevel} and DynaBOA~\cite{guan2022out} adapt by supervising with ground-truth 2D poses and using images from the pre-training dataset as exemplars. 
CycleAdapt~\cite{nam2023cyclic} further extends to practical scenarios by referring to 2D pose detections, and enhances adaptation by employing denoised motion estimations as pseudo labels to capture 3D geometry.

\begin{figure}
\centering
\includegraphics[width=.9\linewidth]{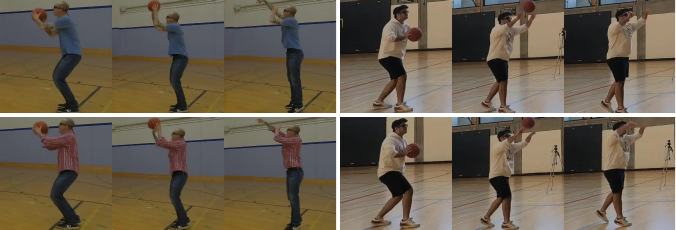}
\vfill
\includegraphics[width=.9\linewidth]{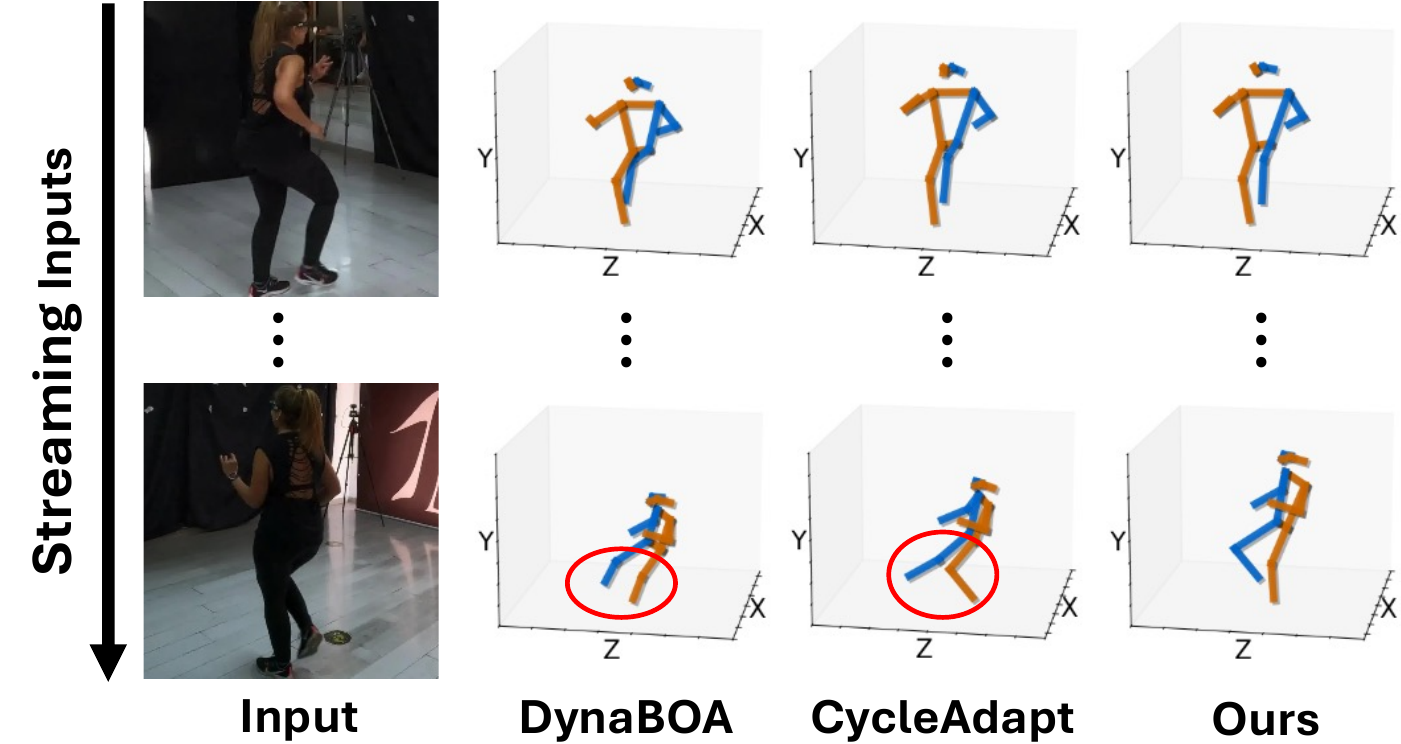}
\caption{Illustration of personal shape and habitual motion traits across observations (upper) and error accumulation in existing works as adaptation progresses (lower).}
\label{fig:personal_traits}
\end{figure}

Nonetheless, as shown in \cref{fig:personal_traits}, we identify two drawbacks in existing works: 
First, these works can be sensitive to self-supervised signals derived from imperfect 2D pose detection and 3D pose estimations, which further leads to error accumulation as prediction errors compound over time, hindering accurate long-term adaptation. 
Second, individuals often exhibit consistent shape and habitual motion traits. 
Model adaptation should benefit from capturing these unique personal patterns when continuously observing and adapting to a single subject. 
However, the struggle with error accumulation positions continuous adaptation as also a risk, leaving personalized adaptation not thoroughly explored.

\begin{figure*}[t]
\centering
\includegraphics[width=0.76\linewidth]{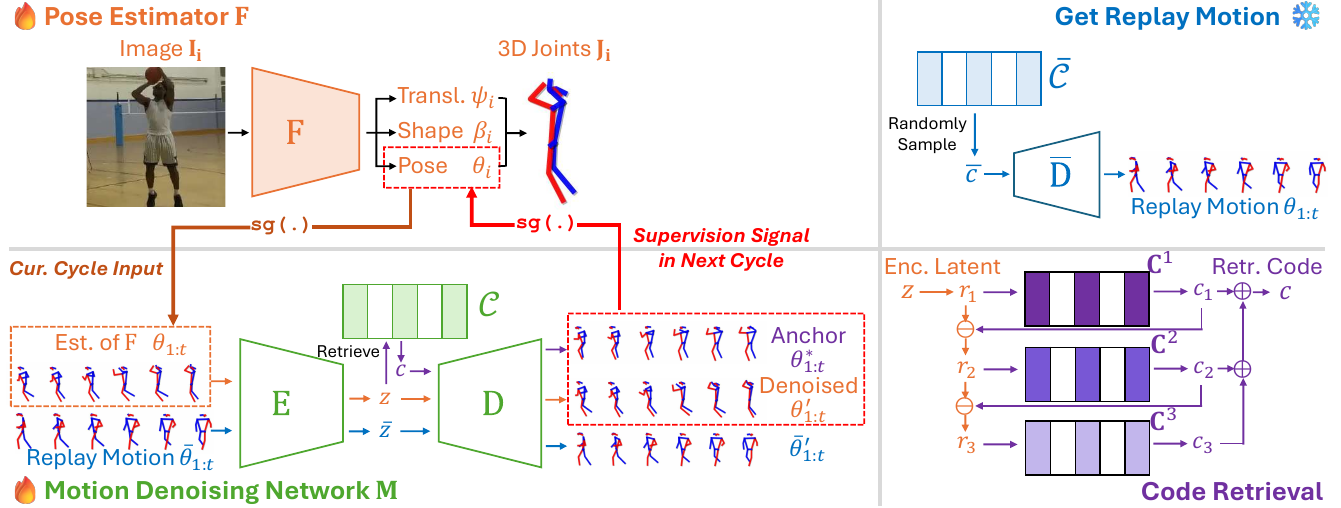}
\caption{Framework Overview. During test time, the pose estimator $\mathrm{F}$ and motion denoising network $\mathrm{M}$ are alternately updated in a cyclic way.
We employ motion discretization to regularize the adaptation of $\mathrm{F}$ and enable self-replay for adapting $\mathrm{M}$.}\label{fig:framework}
\end{figure*}

To address these limitations, we propose a novel solution that integrates motion discretization and a soft-reset strategy, thus mitigating error accumulation in long-term adaptation. 
In this way, our solution enables enhanced online test-time adaptation through continuously observing and adapting to a personalized test domain.
This test domain comprises streaming videos featuring a single individual who is not included in the pre-training dataset, with total recording durations potentially extending to tens of minutes.
As shown in \cref{fig:framework,alg:adaptation}, our framework consists of two components, which are adapted alternately in a cyclic manner~\cite{nam2023cyclic}. 
The first component is a pose estimator that outputs 3D human poses corresponding to the input images. 
The second is an autoencoder-based motion denoising network that refines these estimations to generate 3D signals capturing the dynamics of human movements, which in turn regularize the adaptation of the pose estimator.

Our key innovations are threefold.
First, we derive a discrete set of diverse anchor motions that capture plausible human movements.
To achieve this, we perform unsupervised clustering on the latent space of the motion denoising network during pre-training, yielding a codebook of representative latent features. 
Each codebook entry is decoded into a distinct anchor motion representing a coherent motion pattern.
During test time, these anchor motions provide regular supervision signals for regularizing the adaptation of the image-based pose estimator, thus mitigating the limitations of self-supervision using imperfect 3D estimations.
Second, our motion discretization also enables a self-replay mechanism, where we adapt the motion denoising network using both incoming test-time estimations and pre-trained anchor motions. 
This ensures consistent decoding of regular anchor motions throughout the adaptation (\cref{fig:dcode}), while eliminating the need to access the original pre-training data.
Finally, we employ an efficient soft reset that periodically reverts the pose estimator to its exponential moving average during adaptation, thus reducing the impact of noisy updates while retaining critical test-time traits for robust adaptation.

We evaluate online personalized test-time adaptation on Ego-Exo4D~\cite{grauman2024ego} and 3DPW~\cite{von2018recovering} datasets.
In line with previous test-time adaptation approaches, we pre-train our model on the Human3.6M dataset~\cite{ionescu2013human3}.
Results demonstrate our enhanced effectiveness compared to previous online test-time adaptation methods and verify our key designs. 
Our contributions are summarized as follows: 
\begin{itemize}
\itemsep0em
\item We propose a novel solution for online test-time adaptation in 3D human pose estimation. 
Our solution highlights using motion discretization and integrates a soft-reset mechanism. 
In this way, we mitigate the inherent challenge of error accumulation in long-term self-supervised test-time adaptation.
\item We examine online test-time adaptation by focusing on a personalized test domain.
By mitigating error accumulation, our solution enables average performance gains through continuous adaptation on streaming videos featuring the same person.
\item We demonstrate enhanced accuracy over existing online test-time adaptation methods.
\end{itemize}

\section{Related Works}
\paragraph{3D Human Body Pose Estimation}
Massive learning-based research has advanced 3D human body estimation given monocular RGB observation. 
This is achieved by exploiting the spatial dimensions to output body poses from single images \cite{kanazawa2018end,kolotouros2019learning,moon2020i2l,zhang2021pymaf}, or further extending to video inputs and further harnessing temporal continuity for enhanced robustness \cite{kanazawa2019learning,kocabas2020vibe,choi2021beyond,wei2022capturing,shen2023global,you2023co}.
These solutions primarily scale training data to enhance the generalizability of pre-trained models across various test scenarios, where recent research~\cite{dwivedi2024tokenhmr,goel2023humans} further leverages more powerful backbones such as ViT~\cite{dosovitskiy2020image}.
Additionally, recent studies also explore pose augmentation for enhanced accuracy~\cite{gong2021poseaug,chai2023global}.

In contrast, another branch of solution addresses from the perspective of test-time adaptation, which directly updates the pre-trained model on unlabeled inputs prior to final estimation, thus mitigating the inherent train-test domain gap for enhanced performance ~\cite{joo2021exemplar,mugaludi2021aligning,guan2021bilevel,guan2022out,weng2022domain,nam2023cyclic,lin2025semantics,lin2025online}.
Specifically, when it comes to online test-time adaptation for streaming inputs,
BOA~\cite{guan2021bilevel} and DynaBOA~\cite{guan2022out} use a bilevel adaptation strategy that updates an image-based pose estimator with supervision of ground-truth 2D keypoints and 2D temporal constraints, while source domain images provide 3D exemplar guidance.
However, their inability to capture the 3D regularity of test motions hinders accurate depth estimation.
Noticing the phenomenon of error accumulation, Lin \etal~\shortcite{lin2025online} restore the updated model to its pre-trained weights at the end of each video and reinitialize with representative historical frames.
CycleAdapt~\cite{nam2023cyclic} introduces a motion denoising network to refine the image-based pose estimation. 
The denoised motion provides 3D pseudo labels for adapting the image-based pose estimator, thereby establishing cyclic adaptation between the two modules.
However, its overreliance on imperfect estimations limits its ability to manage error accumulation in long-term adaptation.

Compared to existing research, we address error accumulation in continuous self-supervised adaptation by leveraging motion discretization and soft reset.
This further enables us to harness personal traits throughout continuous adaptation on observations featuring the same person, resulting in enhanced performance.

\paragraph{Quantized Body Pose Representation}
Although human body poses are inherently represented as continuous values in 3D space, recent research has exploited discrete pose representations for both human pose estimation and motion generation. 
On one hand, a trend in human motion generation~\cite{lucas2022posegpt,zhang2023generating, guo2024momask} employs VQ-VAE~\cite{van2017neural} to compress and quantize consecutive pose frames into discrete latent tokens. 
These discrete tokens are then integrated into a GPT-like model, enabling long-term generation of valid and plausible poses through next-token prediction. 
On the other hand, recent research~\cite{geng23PCT,dwivedi2024tokenhmr} applies this quantization philosophy to facilitate image-based human pose estimation, as achieved by learning latent codebooks that capture the spatial relationships among groups of joints.
This reduces pose estimation to a latent code prediction problem, which helps obtain valid outputs that accurately capture body pose physics.

In comparison, our solution emphasizes motion discretization in self-supervised adaptation for human body pose estimation. 
Derived from unsupervised clustering in the latent motion space, our anchor motions model the spatiotemporal relationships among joints by capturing both pre-trained regularity and test-time motion traits, thus mitigating error accumulation during long-term adaptation.

\section{Methodology}

Our online test-time personalized adaptation defines its test domain as a streaming video $\mathcal{S}$, which consists of concatenated image sequences featuring the same individual who was unseen during pre-training. 
During test time, we adapt the model on each incoming unlabeled batch $\mathcal{V} \subset \mathcal{S}$, making on-the-fly prediction of 3D body joints $\vb{J} \in \mathbb{R}^{N\times3}$ for each image $\vb{I} \in \mathcal{V}$.

As shown in \cref{fig:framework}, our framework consists of two components: a pose estimator $\mathrm{F}$ and a motion denoising network $\mathrm{M}$. 
$\mathrm{F}$ maps an input image $\vb{I} \in \mathcal{V}$ to SMPL~\cite{loper2015smpl} pose $\vb*{\theta}$, shape $\vb*{\beta}$, and translation $\vb*{\psi}$, from which the 3D joints $\vb{J}$ are subsequently regressed.
$\mathrm{M}$ is a denoised autoencoder that captures temporal continuity across outputs of $\mathrm{F}$ and generates 3D motion signals to help adapt the $\mathrm{F}$. 
$\mathrm{F}$ and $\mathrm{M}$ are adapted alternately to capture test-time appearance and motion traits in a cyclic manner~\cite{nam2023cyclic}.

We summarize our test-time pipeline in \cref{alg:adaptation} and introduce our key designs in the following subsections. 
We first introduce our motion discretization, which is achieved through unsupervised clustering and benefits test-time adaptation in two ways:
first, it regularizes the adaptation of $\mathrm{F}$ to mitigate the effects of self-supervision with noisy estimations; 
second, it enables a self-replay mechanism when adapting the motion denoising network $\mathrm{M}$, addressing representation drift to ensure consistent decoding of regular anchor motions.
We further introduce our soft-reset strategy and provide implementation details.

\begin{algorithm}[!t]
\caption{Our adaptation pipeline.}
\small
\begin{algorithmic}[1]
\REQUIRE Streaming video $\mathcal{S}$ featuring the test person, pre-trained pose estimator $\bar{\mathrm{F}}$, motion denoising network $\bar{\mathrm{M}} = (\bar{\mathrm{E}},\bar{\mathrm{D}}) $ and codebook $\bar{\mathcal{C}}$
\ENSURE 3D body keypoints $\vb{J}\in\mathbb{R}^{N\times3}$ for every image $\vb{I}\in \mathcal{S}$.
\STATE Initialize $\mathrm{F}, \mathrm{E}, \mathrm{D}, \mathcal{C}$ with $\bar{\mathrm{F}}, \bar{\mathrm{E}}, \bar{\mathrm{D}}, \bar{\mathcal{C}}$, respectively
\WHILE{incoming $\mathcal{V}\subset\mathcal{S}$}
\STATE Set $\mathrm{F}_{pre}\leftarrow \mathrm{F}$. Prepare replay motion $\bar{\vb*{\theta}}_{1:t}$ by \cref{eq:replay}.
\FOR{cycle=0,...,c}
\STATE \emphalg{\# Adapt Pose Estimator $\mathrm{F}$}
\STATE Retrieve $\vb*{\beta}^\prime,\vb*{\theta}^\prime$ and anchor $\vb*{\theta}^\ast$ obtained in previous cycle
\STATE $\vb*{\beta},\vb*{\theta},\vb*{\psi},\vb{J} \leftarrow \mathrm{F}(\vb{I})$ for $\vb{I}\in\mathcal{V}$
\STATE Update $\mathrm{F}$ with $L_F$ (\cref{eq:loss_f}) \commentalg{loss with anchor motion}
\STATE \emphalg{\# Adapt Motion Denoising Network $\mathrm{M}$}
\STATE $\vb*{z}\leftarrow\mathrm{E}(\vb*{\theta}_{1:t}), \bar{\vb*{z}}\leftarrow\mathrm{E}(\bar{\vb*{\theta}}_{1:t})$
\STATE For $\vb*{z}$, retrieve nearest code $\vb*{c}$ from $\mathcal{C}$ (\cref{eq:code}).
\STATE Anchor $\vb*{\theta}^\ast_{1:t}\leftarrow\mathrm{D}(\vb*{c})$, denoised $\vb*{\theta}^\prime_{1:t}\leftarrow\mathrm{D}(\vb*{z})$, $\bar{\vb*{\theta}}^\prime_{1:t}\leftarrow\mathrm{D}(\bar{\vb*{z}})$
\STATE Update $\mathrm{E},\mathrm{D}$ with $L_M$ (\cref{eq:loss_m}), $\mathcal{C}$ with $\bar{\vb*{z}}$ \commentalg{self-replay}
\ENDFOR
\STATE \emphalg{\# Get final prediction}
\STATE $\vb*{\beta},\vb*{\theta},\vb*{\psi},\vb{J}  \leftarrow \mathrm{F}(\vb{I})$
\STATE Soft reset $\mathrm{F}$ by \cref{eq:soft_reset}
\commentalg{soft reset}
\ENDWHILE
\end{algorithmic}\label{alg:adaptation}
\end{algorithm}

\subsection{Unsupervised Clustering for Motion Discretization}\label{sec:clustering}

We perform unsupervised clustering on the latent space of the motion denoising network $\mathrm{M}$, yielding a codebook $\mathcal{C}$ of discrete latent motion representations. 
Inspired by recent work on motion generation~\cite{guo2024momask}, our $\mathcal{C}=\{\vb{C}^i\in\mathbb{R}^{N_c\times d}|i=1,...,k\}$ adopts a residual design with $k$ layers, with each layer having $N_c$ codes of dimension $d$.
The anchor motion set is then obtained by decoding codebook codes, which capture diverse yet regular human movements spanning the motion space. 

We construct $\mathcal{C}$ during the pre-training of $\mathrm{M}$.
While the encoder $\mathrm{E}$ and decoder $\mathrm{D}$ are optimized as a standard denoising autoencoder, $\mathcal{C}$ is updated using the exponential moving average with the encoded latent $\vb*{z}\in\mathbb{R}^d$ of the input motion. 
Specifically, we initialize the residual $\vb*{r}_1=\vb*{z}$. 
For each layer $i$, we recursively retrieve the nearest code $\vb*{c}_i\in\vb{C}^i$ and update the residual $\vb*{r}_{i+1}$:
\begin{equation}
    \vb*{c}_i  = \argmin_{\vb*{c}\in \vb{C}^i}||\vb*{c}-\vb*{r}_i||_2, \vb*{r}_{i+1}=\vb*{r}_i-\vb*{c}_i.\label{eq:code}
\end{equation}
The selected codes $\vb*{c}_i$ are then updated using the corresponding $\vb*{r}_i$ via exponential moving average.

\subsection{Regularize with Motion Discretization to Adapt Pose Estimator}\label{sec:reg_term}

During test time, we obtain anchor motion and use it to regularize the adaptation of $\mathrm{F}$. 
Denoting the output of $\mathrm{F}$ over $t$ frames as $\vb*{\theta}_{1:t}$, we encode it into the latent $\vb*{z} = \mathrm{E}(\vb*{\theta}_{1:t})$. 
$\vb*{z}$ is then quantized to obtain $\vb*{c} = \sum_{i=1}^k \vb*{c}_i$. 
Each $\vb*{c}_i$ is retrieved by \cref{eq:code}, utilizing the latent regularity in measuring motion similarity.
Both original latent $\vb*{z}$ and quantized code $\vb*{c}$ are fed in parallel to the decoder, yielding denoised motion $\vb*{\theta}^\prime_{1:t} = \mathrm{D}(\vb*{z})$ and anchor motion $\vb*{\theta}^\ast_{1:t} = \mathrm{D}(\vb*{c})$.

The anchor motion then supervises the update of $\mathrm{F}$ in the next cycle, with the anchor loss defined as 
\begin{equation}\label{eq:loss_anchor}
L_{ach} = ||\vb*{\theta}-\mathtt{sg}(\vb*{\theta}^\ast)||,
\end{equation}
where $\mathtt{sg}(.)$ denotes the stop-gradient operator. 
Similar to human motion generation works~\cite{lucas2022posegpt,zhang2023generating,guo2024momask} that predict discrete latent motion codes for valid and realistic generation, we empirically find a comparable effect.
While $\vb*{\theta}^\prime_{1:t}$ can be corrupted by self-supervision with noisy estimations (see \cref{eq:loss_m,fig:framework}), the discretized anchor motion $\vb*{\theta}^\ast_{1:t}$ could filter out high-frequency noise while retaining core coherent motion patterns, thus regularizing the network update to facilitate long-term adaptation.

The overall loss for adapting $\mathrm{F}$ integrates $L_{ach}$  with other signals introduced in CycleAdapt~\cite{nam2023cyclic}, \ie, the detected 2D keypoints, the temporally averaged shape $\vb*{\beta}^\prime$, and the denoised motion $\vb*{\theta}^\prime_{1:t}$:
\begin{equation}
L_F = L_{p} + \lambda_1 L_{s} +\lambda_2 L_{2D} + \lambda_3 L_{ach},\label{eq:loss_f}
\end{equation}
where $L_{p} = ||\vb*{\theta}-\mathtt{sg}(\vb*{\theta}^\prime)||, L_{s} = ||\vb*{\beta}-\mathtt{sg}(\vb*{\beta}^\prime)||$, and $L_{2D}$ is the reprojection error that compares the 2D projections of the estimations with the 2D keypoints obtained from off-the-shelf detectors.
$\lambda_1,\lambda_2,\lambda_3$ are weighting parameters.

\subsection{Self-Replay with Motion Discretization to Adapt Motion Denoising Network}\label{sec:replay}

The online adaptation of $\mathrm{M}$ can suffer from representation drift, a phenomenon where latent codes gradually lose their ability to be decoded into consistent and regular anchor motions as adaptation progresses (\cf AQM~\cite[Fig.~1]{caccia2020online}). 
As inspired by AQM~\cite{caccia2020online}, our motion discretization further enables a self-replay mechanism to address this challenge.

To implement this, we construct a batch of replay motions during adaptation.
As shown in \cref{fig:framework}-top right, we first sample a random code $\bar{\vb*{c}}=\sum_{i=1}^k \bar{\vb*{c}}_i$ from the pre-trained $\bar{\mathcal{C}}$, and then decode it with the pre-trained decoder $\bar{\mathrm{D}}$ to obtain the replay motion
\begin{equation}
\bar{\vb*{\theta}}_{1:t}=\bar{\mathrm{D}}(\bar{\vb*{c}}).\label{eq:replay}
\end{equation}

We then feed both replayed motions $\bar{\vb*{\theta}}_{1:t}$ and test-time estimations $\vb*{\theta}_{1:t}$ into $\mathrm{M}$ in parallel, with input frames randomly masked to encourage more robust adaptation. 
$\mathrm{M}$ is then updated in a self-supervised way, with the loss defined as
\begin{equation}
L_M = ||\bar{\vb*{\theta}}^\prime_{1:t}-\mathtt{sg}(\bar{\vb*{\theta}}_{1:t})||+ ||\vb*{\theta}^\prime_{1:t}-\mathtt{sg}(\vb*{\theta}_{1:t})||,\label{eq:loss_m}
\end{equation}
where $\bar{\vb*{\theta}}^\prime_{1:t},\vb*{\theta}^\prime_{1:t}$ are the decoded outputs from the encoded latents $\bar{\vb*{z}},\vb*{z}$ of the replay and test-time motions. 
We further update the codebook $\mathcal{C}$ using the replay latent $\bar{\vb*{z}}$ via an exponential moving average with decay $\mu_\mathcal{C}=0.999$, facilitating the synchronization of $\mathcal{C}$ with the evolving latent space.

In this way, we efficiently distill the pre-trained regularity without accessing the original training samples, thus addressing privacy concerns while enabling consistent and regular anchor motions (see \cref{sec:ablation,fig:dcode}) to aid in supervising the pose estimator $\mathrm{F}$.

\subsection{Soft Reset for the Pose Estimator}\label{sec:soft_reset}
After adapting to each batch $\mathcal{V}$, we further perform a soft reset for the pose estimator $\mathrm{F}$.
Specifically, let $\mathrm{F}_{pre}$ and $\mathrm{F}$ respectively denote the weights before and after adapting on $\mathcal{V}$. 
We then set the weights of the pose estimator for subsequent adaptations as follows:
\begin{equation}
\mathrm{F}\leftarrow \mu_\mathrm{F}  \mathrm{F}_{pre} + (1-\mu_\mathrm{F})\mathrm{F},\label{eq:soft_reset}
\end{equation}
where $\mu_\mathrm{F}$ is the decay factor. 
This soft reset, implemented as an exponential moving average during adaptation, reduces the impact of noisy updates on individual batches while retaining critical traits learned from historical adaptation.

\begin{table*}[!t]
\small
\centering
\resizebox{0.99\linewidth}{!}{
\begin{tabular}{c|cc|cc|cc|cc|ccc}
\hline
& \multicolumn{8}{c|}{Ego-Exo4D} &  \multicolumn{3}{c}{\multirow{2}{*}{3DPW}}\\
\cline{2-9}
&   \multicolumn{2}{c|}{Basketball} &  \multicolumn{2}{c|}{Soccer} &  \multicolumn{2}{c|}{Dance}  &  \multicolumn{2}{c|}{All Scenarios} &&&\\
\cline{2-12}
&   MPJPE & \makecell{MPJPE-PA} &  MPJPE & \makecell{MPJPE-PA}&  MPJPE &\makecell{MPJPE-PA} &  MPJPE &\makecell{MPJPE-PA} & MPJPE &\makecell{MPJPE-PA} & MPVPE \\
\hline
Pre-trained $\mathrm{F}$ & 215.8 & 122.7 & 198.6 & 114.7 & 198.1 & 111.0 & 205.8 & 116.5 & 230.3 & 123.4 & 253.4\\
\hline
\multicolumn{12}{c}{w/ OpenPose~\cite{cao2017realtime}} \\
\hline
BOA~\cite{guan2021bilevel}$^\dag$ & 158.4 & 72.8 & 115.3 & 63.3 & 117.8 & 69.3 & 135.1 & 70.0 & 98.2 & 55.8 & 114.2 \\
DynaBOA~\cite{guan2022out}$^\dag$ & 173.0 & 73.4 & 129.1 & 68.2 & 141.0 & 70.9 & 153.3 & 71.6 & 139.7 & 63.8 & 155.1 \\
CycleAdapt~\cite{nam2023cyclic} & 160.4 & 80.8 & 126.3 & 78.5 & 135.3 & 81.0 & 145.0 & 80.5 &  141.0 & 79.6 & 155.6 \\
Ours & \textbf{141.6} & \textbf{71.3} & \textbf{105.3} & \textbf{61.3}  & \textbf{106.4} & \textbf{67.0} & \textbf{121.5} & \textbf{68.1} & \textbf{83.9} & \textbf{51.6} & \textbf{100.3}\\
\hline
\multicolumn{12}{c}{w/ ViTPose~\cite{xu2022vitpose+,xu2022vitpose}} \\
\hline
BOA~\cite{guan2021bilevel}$^\dag$ & 149.6 & 65.3 & 108.9 & 58.2 & 101.9 & 61.9 & 123.5 & 62.9 &  91.5 & 53.9 & 105.5\\
DynaBOA~\cite{guan2022out}$^\dag$ & 154.8 & 64.3 & 114.8 & 58.6 & 108.5 & 61.3 & 129.4 & 62.2 & 118.4 & 56.1 & 134.5\\
CycleAdapt~\cite{nam2023cyclic} & 144.8 & 69.6 & 112.1 & 62.9 & 108.3 & 67.1 & 124.6 & 67.6 & 109.6 & 66.6 & 124.8 \\
Ours &  \textbf{137.4} & \textbf{63.4} & \textbf{102.2} & \textbf{56.8} & \textbf{99.8} & \textbf{59.5} & \textbf{116.4} & \textbf{60.8} & \textbf{85.0} & \textbf{53.3} & \textbf{100.4} \\
\hline
\end{tabular}}
\caption{Comparison with the pre-trained pose estimator and existing online test-time adaptation methods on Ego-Exo4D and 3DPW. $\dag$ denotes leveraging original pre-trained data in adaptation.}\label{tab:online_egoexo}\label{tab:online_3dpw}
\end{table*}

\subsection{Implementation Details}\label{sec:implementation}
We include the network architecture and pre-training details in the appendix.
We adopt the same pre-trained pose estimator $\mathrm{F}$ from CycleAdapt~\cite{nam2023cyclic}, which builds on a ResNet-50~\cite{he2016deep} backbone.
Our $\mathrm{M}$ processes inputs and outputs consisting of $t=16$ frames at 15 fps, with a latent dimension of $d=512$.
The residual codebook $\mathcal{C}$ has $k=3$ layers, each with $N_c=512$ codes.

During test time, we segment $\mathcal{S}$ into sequential batches $\mathcal{V}$, each containing 160 frames at 30 fps. 
Most of the hyperparameters are adopted from the implementation of CycleAdapt~\cite{nam2023cyclic}.
For each $\mathcal{V}$, we use the Adam optimizer~\cite{kingma2014adam} with an initial learning rate of $5\times 10^{-5}$, which is further reduced to $1\times 10^{-6}$ using a cosine annealing~\cite{loshchilov2016sgdr}. 
The number of cycles is set to $c=12$, with a mini-batch size of 32 for both $\mathrm{F},\mathrm{M}$ within each cycle.
For $\mathrm{F}$, we set $\lambda_1 = 0.001, \lambda_2 = 0.1, \lambda_3 = 0.3$ to balance different loss terms, and set $\mu_F=0.95$ for our soft-reset strategy.
When adapting $\mathrm{M}$, we use a mini-batch size of 4 for replay motion.

\section{Experiment}

\subsection{Dataset and Evaluation Protocol}\label{sec:protocol}

\paragraph{Human3.6M} is a large-scale dataset collected in controlled indoor settings~\cite{ionescu2013human3}. 
We follow previous test-time adaptation approaches~\cite{guan2021bilevel,guan2022out,nam2023cyclic} to pre-train ours on this dataset.

\paragraph{Ego-Exo4D} has skilled human activities collected from cities worldwide~\cite{grauman2024ego}. 
We focus on basketball, soccer, and dancing scenarios, which present diverse body motion dynamics and action repetitions for capturing personal traits.
Adapting to this dataset poses challenges due to the domain gap introduced by fisheye camera recordings and various motions.
For evaluation, we select 30 participants and utilize their undistorted exocentric videos.
The total recording durations of each participant range from 10 to 50 minutes. 
Detailed statistics are provided in the appendix.

We report the average performance over participants by calculating the \textbf{Mean Per Joint Position Error (MPJPE)} in the root-aligned space and the \textbf{Procrustes-Aligned MPJPE (MPJPE-PA)} in millimeters (\texttt{mm}), across 17 keypoints defined in MS COCO format~\cite{lin2014microsoft}. 
We did not evaluate mesh reconstruction due to the lack of annotations.

\begin{figure}[!t]
\centering
\includegraphics[width=.99\linewidth]{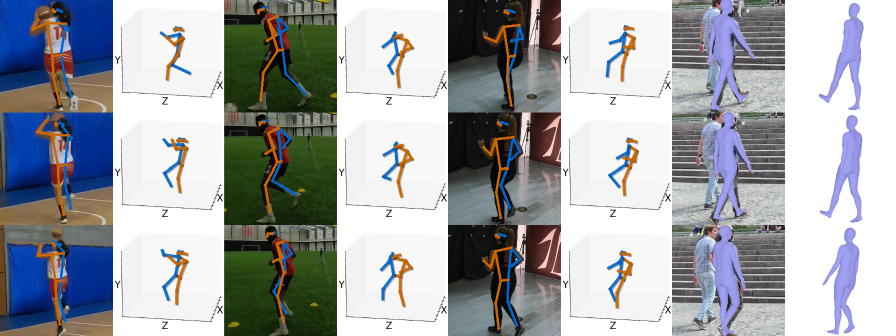}
\caption{Our qualitative results. 
We show both 2D projections and 3D estimations in camera space.}\label{fig:quali_cases}
\end{figure}

\begin{table}[!t]
\small
\centering
\resizebox{0.99\linewidth}{!}{
\begin{tabular}{>{\centering\arraybackslash}m{4.5cm}|cc|ccc}
\midrule
&  \multicolumn{2}{c|}{Ego-Exo4D} & \multicolumn{3}{c}{3DPW}\\
\cline{2-6}
& MPJPE & \makecell{MPJPE\\-PA} &  MPJPE & \makecell{MPJPE\\-PA}&  MPVPE \\
\hline
Pre-trained $\mathrm{F}$ & 205.8 & 116.5 & 230.3 & 123.4 & 253.4\\
\hline
SPIN~\cite{kolotouros2019learning} & 137.3 & 71.5 & 96.9 & 59.2 & 116.4 \\
I2L-Mesh~\cite{moon2020i2l} & 136.6 & 73.8 & 93.2 & 57.7 &110.1\\
PyMAF~\cite{zhang2021pymaf} & 135.5 & 70.2 & 92.8 & 58.9 & 110.1 \\
HMR-2.0b~\cite{goel2023humans} $\dag$ & 125.2 & 65.6 & \textbf{81.3} & 54.3 & \textbf{94.4} \\
\hline
VIBE~\cite{kocabas2020vibe}  & 130.6 & 70.1 & 93.5 & 56.5 & 113.4 \\
TCMR~\cite{choi2021beyond} & 126.4 & 66.9 & 95.0 & 55.8 & 111.3 \\
GLoT~\cite{shen2023global} & 127.1 & 66.7 & 89.9 & 53.5 & 107.8 \\
\hline
Ours (w/ OpenPose) & 121.5 & 68.1 & 83.9 & \textbf{51.6} & 100.3 \\
Ours (w/ ViTPose) & \textbf{116.4} & \textbf{60.8} & 85.0 & 53.3 & 100.4 \\
\hline
\end{tabular}}
\caption{Comparison with domain generalization methods that leverage more training datasets.
All compared methods do not leverage Ego-Exo4D and 3DPW for pre-training. 
$\dag$ indicates using a ViT-H/16 backbone.}\label{tab:dg}
\end{table}

\begin{figure*}[!t]
\centering
\includegraphics[width=.88\linewidth]{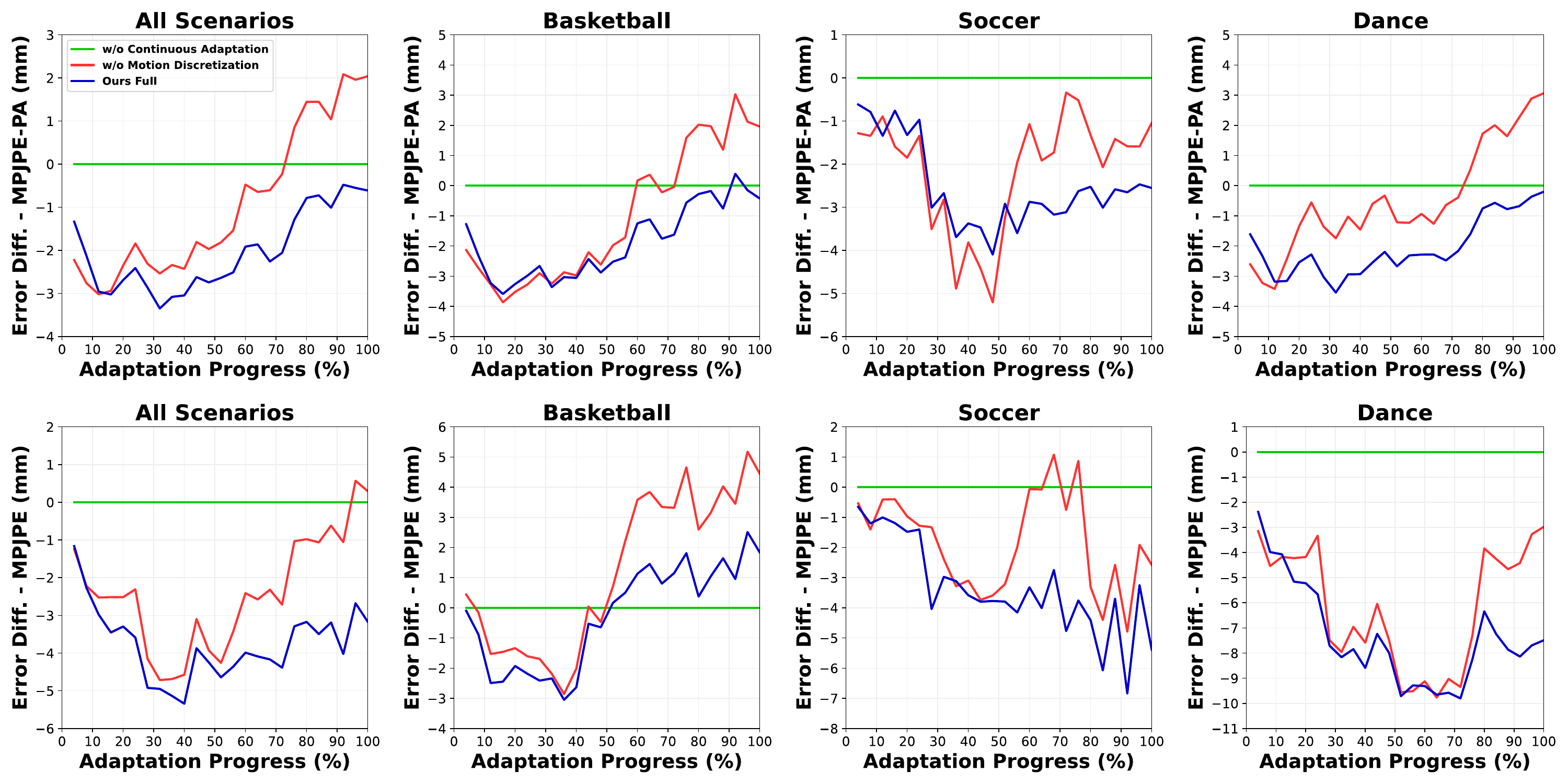}
\caption{Error difference versus adaptation progress over time, with adaptation progress expressed relative to the total recording length for each participant.
We plot the error difference relative to the baseline \textit{w/o continuous adaptation} ({\color[rgb]{0.,0.8,0.}{green}}), which always starts from pre-trained weights for each $\mathcal{V}$. 
Our complete solution ({\color[rgb]{0.,0.,0.8}{blue}}) is compared against the variant of \textit{w/o motion discretization} ({\color[rgb]{1.,0.2,0.2}{red}}), which removes both $L_{ach}$ and self-replay. 
A lower $y$-axis value indicates better performance.}\label{fig:error_vs_time}
\end{figure*}

\paragraph{3DPW} is an in-the-wild dataset collected with a moving hand-held camera, which has recordings with various appearances and backgrounds for each participant~\cite{von2018recovering}.
We follow previous test-time adaptation studies~\cite{guan2021bilevel,guan2022out,nam2023cyclic} to evaluate on its test split, which includes 5 participants with a total number of 35,515 frames at 30 fps.
For adaptation, we set the personalized test domain.
We evaluate the estimated joints with \textbf{MPJPE} and \textbf{MPJPE-PA}, and further compare the output mesh with the ground truth to report the \textbf{Mean Per Vertex Position Error (MPVPE)} in \texttt{mm}.

\subsection{Comparison with Related Works}\label{sec:comparison}

\paragraph{Discussion on Online Test-time Adaptation} 
We first focus on comparison with existing online test-time adaptation research that has publicly available code, including BOA~\cite{guan2021bilevel}, DynaBOA~\cite{guan2022out}, and CycleAdapt~\cite{nam2023cyclic}. 
All compared methods are pre-trained on the Human3.6M dataset, employ a ResNet-50 backbone for the image-based pose estimator, adapt to the same video stream, and utilize the same 2D detections obtained by either OpenPose~\cite{cao2017realtime} or ViTPose~\cite{xu2022vitpose,xu2022vitpose+}, ensuring a fair comparison.
We also include the pre-trained pose estimator $\mathrm{F}$ as a baseline, which is directly applied to the test data using its pre-trained weights.

In~\cref{tab:online_egoexo,tab:online_3dpw}, we respectively report quantitative comparisons on Ego-Exo4D and 3DPW.
We show our qualitative results in \cref{fig:quali_cases}, and provide qualitative comparisons with previous works in the appendix. 
Together, these results demonstrate our superior performance compared to existing online test-time adaptation methods.
This is achieved because our designs mitigate challenges in long-term adaptation, allowing us to effectively exploit personal traits for robust estimation.
Furthermore, we notice that using 2D detections from ViTPose can enhance our performance on Ego-Exo4D but offers limited improvement on 3DPW. 
We attribute this to the observation that 3DPW often contains overlapping participants, which ViTPose struggles to separate effectively.

In comparison, we observe that DynaBOA and CycleAdapt suffer from severe error accumulation under self-supervision with imperfect estimations, resulting in decreased accuracy through continuous adaptation. 
Furthermore, BOA and DynaBOA report larger errors, especially in terms of MPJPE and MPVPE. 
This is primarily due to their heavy reliance on erroneous 2D supervision without adequate 3D guidance, leading to inaccurate depth estimation, as aligned with the discussion in \cite{nam2023cyclic}.

We also discuss the run-time cost in the appendix, demonstrating competitive efficiency offered by our proposed solution.
In contrast, BOA and DynaBOA require storing pre-trained images for exemplar guidance, which increases storage demands and run-time due to the exemplar retrieval.

\paragraph{Discussion with Domain Generalization Methods}
In \cref{tab:dg}, we conduct an empirical comparison with domain generalization methods that directly apply the pre-trained model to the test domain. 
These baselines pre-train on extensive datasets beyond Human3.6M, where HMR-2.0~\cite{goel2023humans} further employs more powerful ViT backbones.

Compared to the reported baselines that utilize a ResNet-50 backbone, our solution demonstrates better performance when using ViTPose detections. 
Additionally, with OpenPose detections, our solution outperforms these baselines on the 3DPW dataset and achieves a lower MPJPE on the Ego-Exo4D.
Compared to HMR-2.0, our solution shows better accuracy on Ego-Exo4D when using 2D detections obtained from ViTPose, and achieves competitive results on 3DPW. 
These comparisons highlight the challenge of applying a pre-trained pose estimator to test domains that differ significantly from the pre-trained ones, while demonstrating the effectiveness of our solution in bridging this train-test domain gap.
We further refer the readers to the appendix for a more detailed discussion.

\begin{table}
\centering
\resizebox{.99\linewidth}{!}{
\begin{tabular}{c|cc|cc}
\hline
\multirow{2}{*}{\makecell{w/ Soft-Reset \\on $\mathrm{F}$}} &  \multicolumn{2}{c|}{w/ Motion Discretization} & \multirow{2}{*}{MPJPE} & \multirow{2}{*}{MPJPE-PA}\\
\cline{2-3}
& \makecell{w/ $L_{ach}$ ($\theta^\ast$)} & \makecell{w/ Self-Replay} & & \\
\hline
& & & 144.3 & 80.6 \\
$\checkmark$ & & & 122.9 & 69.2 \\
& $\checkmark $ & $\checkmark$ & 138.2 & 74.8 \\
$\checkmark$ & $\checkmark $ & $\checkmark$  & \textbf{121.5} & \textbf{68.1} \\
\hline
$\checkmark$ & & & 122.9 & 69.2 \\
$\checkmark$ &  $\checkmark$ & &  123.2 & 69.2 \\
$\checkmark$ &  & $\checkmark$ &  122.9 & 69.2 \\
$\checkmark$ & $\checkmark $ & $\checkmark$  & \textbf{121.5} & \textbf{68.1} \\
\hline

\hline
\end{tabular}}
\caption{Ablation study and discussion of key components on Ego-Exo4D. In \cref{fig:error_vs_time}, we further demonstrate that motion discretization benefits long-term adaptation.}\label{tab:main_ablation}
\end{table}

\begin{table*}
\centering
\begin{subtable}{.35\textwidth}
\small
\centering
\begin{tabular}{c|c|cc}
\hline
\makecell{w/ $L_p$ \\($\theta^\prime$)} &\makecell{w/ $L_{ach}$\\ ($\theta^\ast)$} &  MPJPE & MPJPE-PA \\
\hline
$\checkmark$ & & 122.9 & 69.2 \\
& $\checkmark$ &  123.1 & 69.8 \\
$\checkmark$ & $\checkmark$ & \textbf{121.5} & \textbf{68.1} \\
\hline
\end{tabular}
\end{subtable}
\hfill
\begin{subtable}{0.28\textwidth}
\centering
\resizebox{.99\linewidth}{!}{
\begin{tabular}{c|cc}
\hline
\makecell{Soft-Reset \\ Decay $\mu_F$} &  MPJPE & MPJPE-PA \\
\hline
0 & 138.2 & 74.8 \\
0.9 & 122.7 & 69.2 \\
0.95 & \textbf{121.5} & \textbf{68.1} \\
1.00 & 125.3 & 70.2 \\
\hline
\end{tabular}}
\end{subtable}
\hfill
\begin{subtable}{0.27\textwidth}
\centering
\resizebox{.99\linewidth}{!}{
\begin{tabular}{c|cc}
\hline
\makecell{w/ Cont. \\ Adapt.}&      MPJPE & MPJPE-PA \\
\hline
    & 125.3 & 70.2 \\
$\checkmark$  & \textbf{121.5} & \textbf{68.1} \\
\hline
\end{tabular}}
\end{subtable}
\caption{Analysis of 3D motion supervision signals, soft-reset decay and continuous adaptation on Ego-Exo4D.}\label{tab:main_ablation2}
\end{table*}

\subsection{Ablation Study and Discussions}\label{sec:ablation}
We further examine our key designs on Ego-Exo4D, where participants have long recordings. 
For all compared methods, we use 2D detections obtained from OpenPose. 
Due to limited space, we report the overall results across all participants here and provide per-scenario results in the appendix.

\begin{figure}[!tb]
\centering
\includegraphics[width=.99\linewidth]{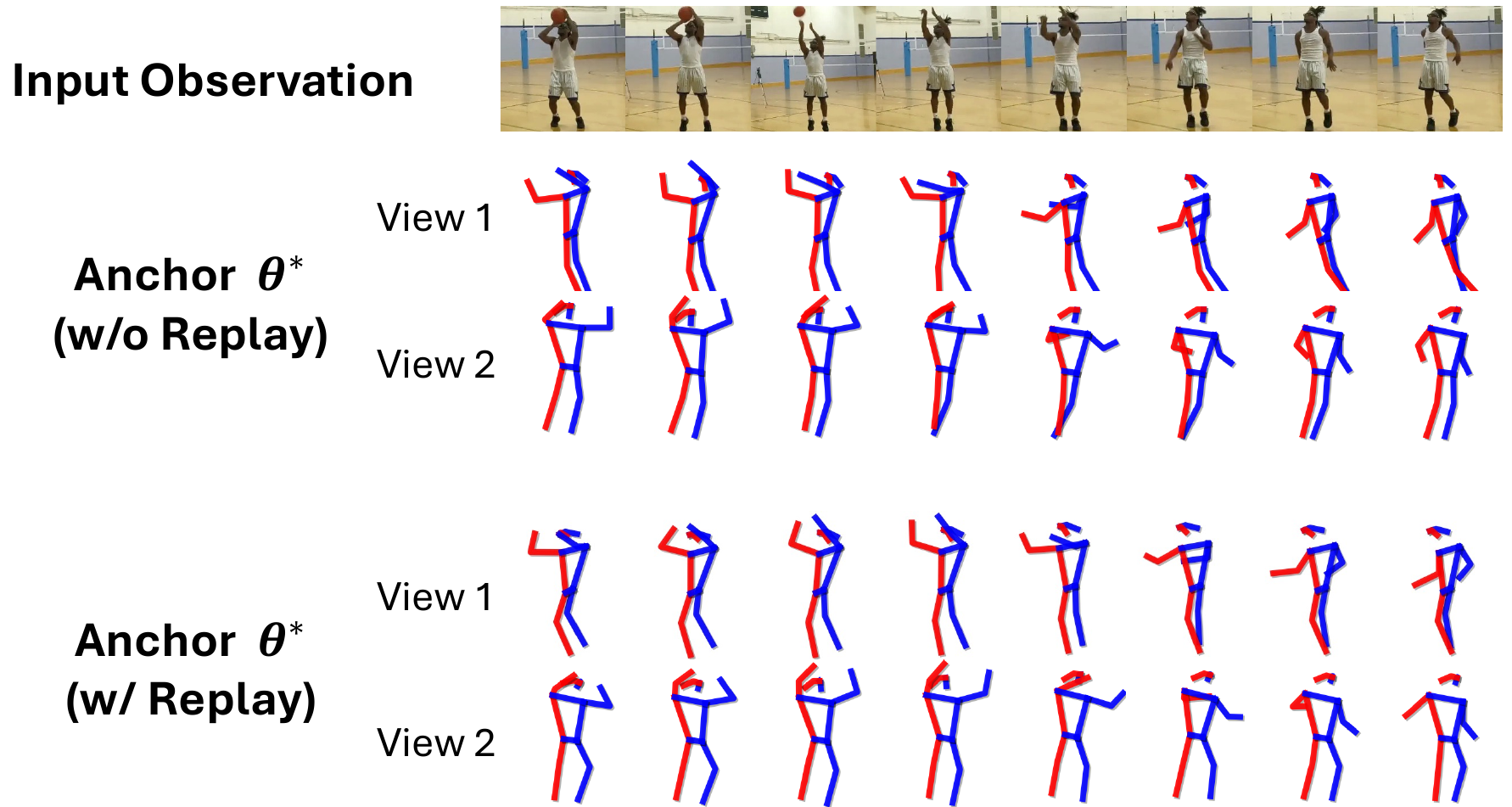}
\caption{Visualization of the retrieved anchor $\vb*{\theta}^\ast$ for the input video depicting basketball shooting, after adapting over \SI{40}{mins} of observation.
Our self-replay mechanism facilitates decoding of realistic and regular anchor motions (\eg leg pose) throughout the adaptation process. 
}\label{fig:dcode}
\end{figure}

\paragraph{Discussion on Motion Discretization}
In \cref{fig:error_vs_time} and \cref{tab:main_ablation}, we demonstrate the benefits of motion discretization in long-term adaptation.
We further find that leveraging motion discretization for both self-replay and anchor loss (\ie, $L_{ach}$ in \cref{eq:loss_anchor}) synergistically achieves optimal performance, as reported in the lower part of \cref{tab:main_ablation}.
This is achieved because our self-replay addresses representation drift to enable decoding regular and realistic anchor motion throughout adaptation.
These anchor motions could then provide faithful references to mitigate the negative influence of self-supervision with imperfect estimations. 
Conversely, using $L_{ach}$ alone increases error due to noisy anchors  (see \cref{fig:dcode}).

In \cref{tab:main_ablation2}-left, we further examine the 3D motion supervision signal in the adaptation of $\mathrm{F}$. 
In addition to the anchor motions $\vb*{\theta}^\ast$ derived from the codebook, we observe that the denoised motion $\vb*{\theta}^\prime$, obtained by decoding the encoded latent $\vb*{z}$, is also beneficial (\ie, $L_p$ in \cref{eq:loss_f}).
While the anchor motions $\vb*{\theta}^\ast$ better preserve regular patterns under long-term self-supervision, $\vb*{\theta}^\prime$ captures fine-grained details that may be lost in discretization. 
Consequently, their complementary roles enable robust and accurate estimation.

Moreover, in the appendix, we demonstrate the benefits of using a residual codebook and updating the codebook with replayed latent $\bar{\vb*{z}}$ when adapting $\mathrm{M}$.
We also compare our motion discretization with an alternative solution that soft resets $\mathrm{M}$ as well, which further verifies the effectiveness of our motion discretization.

\paragraph{Discussion on Soft Reset}
In \cref{tab:main_ablation}-upper, we demonstrate the benefits of the proposed soft reset, where we further discuss the decay $\mu_F$ in \cref{tab:main_ablation2}-middle.
Our soft reset enhances the performance by leveraging test-time traits captured from historical adaptation while mitigating the effects of noisy adaptation on individual batches.
This is evident when compared with two baselines: $\mu_F=0$, which updates solely based on weights from previous adaptation, and $\mu_F=1$, which always resets $\mathrm{F}$ to its pre-trained weights after adaptation on each batch $\mathcal{V}$. 
Moreover, we empirically find that a conservative choice of $\mu_F=0.95$ is beneficial.

\paragraph{Discussion on Continuous Personalized Adaptation}
As reported in \cref{fig:error_vs_time} and \cref{tab:main_ablation2}-right, we achieve enhanced performance on average by exploiting personalized traits through continuous adaptation. 
Here, we set the baseline as resetting the model to its pre-trained weights before adapting to each incoming batch $\mathcal{V}$. 

Specifically, as shown in \cref{fig:error_vs_time}, the performance gain from continuous adaptation is evident in soccer and dancing scenarios.
However, we observe limited benefits for long-term adaptation in basketball scenarios, as some participants are often severely distorted in videos due to the camera setups.
This further enlarges domain gaps and leads to significant batch-wise errors, presenting difficulties for continuous adaptation. 
Detailed analysis is provided in the appendix.

\section{Conclusion}\label{sec:conclusion}
We propose a novel solution to address the challenges inherent in long-term online adaptation with self-supervision, which leverages motion discretization obtained through unsupervised clustering and incorporates an efficient soft-reset mechanism.
Specifically, our motion discretization not only regularizes the adaptation process but also supports a self-replay mechanism. 
We further examine online test-time adaptation on a personalized domain, where our solution enables harnessing the benefits of continuous personalized adaptation and outperforms existing online test-time adaptation methods.
Evaluations demonstrate our effectiveness and verify our key designs.

\paragraph{Limitations and Future Work} 
We adopt a pose estimator with ResNet-50 backbones to output poses in local frames, aligning with previous test-time adaptation methods to illustrate the challenge of error accumulation and demonstrate the effectiveness of our solution.
Nonetheless, we acknowledge recent advancements in developing more powerful estimators using ViT backbones~\cite{goel2023humans} and in recovering global human trajectories in the world coordinate system from a moving camera~\cite{shin2024wham}. 
We also notice advanced techniques that could better exploit test-time traits while keeping stable adaptation, such as adaptive parameters for loss weights or EMA schedules.
We consider integrating these advancements as potential extensions to our solution.
Additionally, we recognize our current limitations in handling participants with distorted appearances, such as those encountered in the basketball scenarios of the Ego-Exo4D dataset. 
We further consider robust adaptation to more challenging egocentric views with severe appearance distortion and self-occlusion as important future work.

\section*{Acknowledgements} 

This research is supported by JSPS KAKENHI Grant Number JP25K03134, Toyota Foundation Grant Number D24-ST-0030, and The Telecommunications Advancement Foundation.

\appendix
\section{Overview}
We divide our appendix into several sections to provide additional implementation details and discussions, thereby supplementing our main text:

In \cref{sec:supp_network}, we present our network architecture and provide pre-training details.

In \cref{sec:supp_protocol}, we offer details regarding our evaluation protocol for the Ego-Exo4D dataset.

In \cref{sec:supp_comparison}, we expand our discussion on comparisons with related work.
We first conduct an empirical comparison with additional domain generalization methods. 
We further provide qualitative and run-time analyses compared to existing online test-time adaptation methods.

In \cref{sec:supp_ablation}, we further examine our design to supplement the discussion in the main text. 
We first provide the distributional analysis and statistical test to examine continuous adaptation and motion discretization.
We then analyze the challenges and performance in basketball scenarios of the Ego-Exo4D dataset. 
Additionally, we present per-scenario results of the ablation studies discussed in the main text and provide a more detailed analysis.

\paragraph{Supplementary Video}
We provide a supplementary video to discuss qualitative comparison and anchor motion, which could be played by most players. 
In our discussion on anchor motion, we demonstrate that the proposed self-replay mechanism facilitates the decoding of regular anchor motions, where the retrieved anchor motions could capture the coherent patterns of input motion throughout adaptation.
Additionally, we visualize the diverse yet regular anchor motions learned during pre-training, where the residual codebook captures the motion patterns from coarse to fine detail.

\section{Network Architecture and Pre-training Details}\label{sec:supp_network}

\subsection{Pose Estimator} 

Our pre-trained pose estimator $\mathrm{F}$ adopts the same architecture and pre-trained weights as the HMRNet of CycleAdapt~\cite{nam2023cyclic}. 
Specifically, $\mathrm{F}$ uses a ResNet-50~\cite{he2016deep} backbone. 
Given an input image $\vb{I}$, $\mathrm{F}$ extracts the image feature from the last layer of the ResNet-50 after average pooling. 
Following HMR~\cite{kanazawa2018end}, $\mathrm{F}$ then uses three fully connected layers to respectively regress the SMPL shape $\vb*{\beta}$, pose $\vb*{\theta}$, and translation $\vb*{\psi}$ from the image feature.

$\mathrm{F}$ is pre-trained on Human3.6M~\cite{goel2023humans}. 
The pre-training follows the scheme of SPIN~\cite{kolotouros2019learning} and leverages the available 3D GT annotations.

\subsection{Motion Denoising Network} 

Our motion denoising network $\mathrm{M}$ is implemented and pre-trained as a denoising autoencoder. 
During its pre-training, we perform unsupervised clustering on the latent space to obtain the codebook $\mathcal{C}$. 
Next, we introduce the implementation details of $\mathrm{M}$, including the input and output motion representations, network architecture, and pre-training details.

\paragraph{Input and Output Motion Representations}
Given the input SMPL parameters for $t$ consecutive frames, we first normalize and convert them into the motion representation $\vb*{\phi}_{1:t}$ before processing with $\mathrm{M}$. 
Specifically, we draw inspiration from STMC~\cite{petrovich2024multi} to construct $\vb*{\phi}_i$ for each frame $i$ as the concatenation of $\vb*{\phi}_i=(y_{root}, \omega_{root,Y}, \vb*{\theta}_+, \vb*{j}_+) \in \mathbb{R}^{d_M}$. 
Here, the positive $y$-axis is denoted as the upward direction, and $y_{root}$ represents the $y$-coordinate (\ie height) of the pelvis.
$\omega_{root,Y}$ is the angular velocity around the $y$-axis of the pelvis, which is computed by comparing neighboring frames.
$\vb*{\theta}_+$ is a 1D vector that concatenates the 6D representation~\cite{zhou2019continuity} of SMPL pose rotations, with the pelvis rotation further normalized by removing its rotation component around the $y$-axis.
$\vb*{j}_+$ is a 1D vector that concatenates root-aligned positions for non-root joints in Euclidean space. 
Consequently, $d_M = 197$ for 22 SMPL joints.

The decoder $\mathrm{D}$ of $\mathrm{M}$ outputs $\vb*{\phi}^\prime_{1:t}=(y^\prime_{root}, \omega^\prime_{root,Y}, \vb*{\theta}^\prime_+, \vb*{j}^\prime_+)_{1:t}$, from which we recover $\vb*{\theta}^\prime_{1:t}$. 
Specifically, we first recover the pelvis's rotation around the $y$-axis by accumulating the output $\omega_{root,Y}^\prime$, starting from the input pelvis rotation of the first frame.
We then obtain the output SMPL pose $\vb*{\theta}^\prime_{1:t}$ by integrating this rotation with the output $\vb*{\theta}^\prime_+$.

\begin{figure*}[!tb]
\centering
\includegraphics[width=0.85\linewidth]{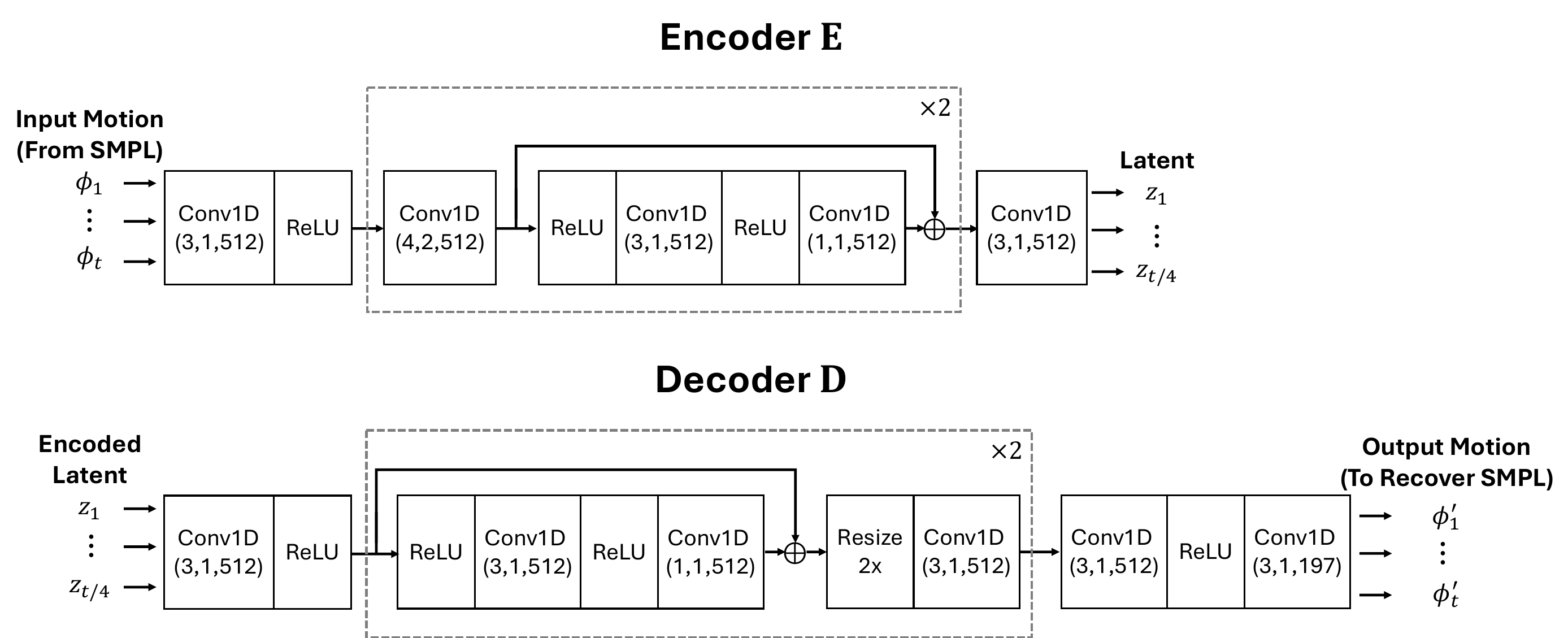}
\caption{Detailed architecture of the encoder $\mathrm{E}$ and decoder $\mathrm{D}$ for our motion denoising network $\mathrm{M}$. 
For 1D convolution blocks, we denote the parameters in the format \texttt{Conv1D(kernel\_size, stride, out\_channels)}}\label{fig:motion_net}
\end{figure*}

\paragraph{Network Architecture} 
$\mathrm{M}$ consists of three components: the encoder $\mathrm{E}$, the decoder $\mathrm{D}$, and the codebook $\mathcal{C}$. 
As mentioned in \cref{sec:clustering} of the main text, we adopt a residual codebook design for $\mathcal{C}$.
For $\mathrm{E},\mathrm{D}$, we visualize their architecture in \cref{fig:motion_net}. 
Following T2M-GPT~\cite{zhang2023generating} and MoMask~\cite{guo2024momask}, both $\mathrm{E}$ and $\mathrm{D}$ consist of two blocks, each utilizing a residual design with 1D convolution and ReLU activation. 
Moreover, each block in $\mathrm{E}$ downsamples along the temporal dimension with a stride of 2, thereby compressing the input of $t$ frame poses into $t/(2^2)$ latent vectors.
$\mathrm{D}$ further takes these compressed latents as input, where each block performs upsampling using nearest-neighbor interpolation, thus recovering frame-wise poses for the original $t$ frames.

\paragraph{Pre-training Details}
We pre-train $\mathrm{M}$ on the Human3.6M dataset by leveraging the SMPL annotations provided by NeuralAnnot~\cite{moon2022neuralannot}, where we further augment the training set by flipping the annotations along the $y$-axis (\ie, left and right).

Given a GT SMPL pose sequence, we further augment it to derive the network input $\vb*{\phi}_{1:t}$ by adding random noise and applying random masks. 
Specifically, we add Gaussian noise $\mathcal{N}(0, \SI{0.015}{rad})$ to the SMPL pose $\vb*{\theta}_{1:t}$ and $\mathcal{N}(0, \SI{0.015}{m})$ to the SMPL translation $\vb*{\delta}_{1:t}$.
We further randomly mask input frames with a probability of 25\%.
The augmented input $\vb*{\phi}_{1:t}$ is encoded by $\mathrm{E}$ to obtain the latent $\vb*{z}$, which is subsequently fed into the decoder $\mathrm{D}$ to produce $\vb*{\phi}^\prime_{1:t}$.

We supervise the learning of $\mathrm{E},\mathrm{D}$ by aligning the output $\vb*{\phi}^\prime=\mathrm{D}(\vb*{z})$ with its GT counterpart $\vb*{\phi}_{GT}$ and using a L1-smooth reconstruction loss. 
This approach enables test-time adaptation to be supervised with also $\vb*{\phi}^\prime$, whose benefits are demonstrated in \cref{tab:main_ablation2}.
We use the AdamW optimizer~\cite{loshchilov2017decoupled} with $\beta_1=0.9$ and $\beta_2=0.99$. 
We adopt an initial learning rate of $2\times 10^{-4}$, which is further decayed to $1\times 10^{-5}$ after 60,000 batches (\ie, in the $360^{\text{th}}$ epoch).
Meanwhile, as described in \cref{sec:clustering} of the main text, for each batch, we use the encoded latent $\vb*{z}$ to update the codebook $\mathcal{C}$ through an exponential moving average, where we set the decay factor as $\mu_{pt}=0.99$.
In total, we pre-train $\mathrm{M}$ for 700 epochs with a batch size of 4096, utilizing an Nvidia RTX 3080Ti GPU with 12GB of memory.

\section{Evaluation Protocol on Ego-Exo4D}\label{sec:supp_protocol}

\begin{table}[t]
\small
\centering
\begin{subtable}{1.\linewidth}
\centering
\begin{tabular}{c|ccccc}
\hline
\texttt{UID}  & 386 & 382 & 387 & 384 & 383 \\
Duration (\texttt{min})  & 48.59 & 47.79 & 46.47 & 45.42 & 37.21\\
\hline
\texttt{UID} &  388& 323 & 318  & 376 & 320  \\
Duration (\texttt{min})  & 36.80& 25.55 & 25.00  & 24.21 & 22.71\\
\hline
\texttt{UID}  & 316  & 558 & 317 \\
Duration (\texttt{min})   & 22.37 & 24.32 & 10.87\\
\hline
\end{tabular}
\caption{Basketball}
\end{subtable}
\vfill
\begin{subtable}{1.\linewidth}
\centering
\begin{tabular}{c|cccc}
\hline
\texttt{UID} & 161 & 549 & 525 & 544 \\
Duration (\texttt{min}) & 15.71 & 14.28 & 11.80 & 10.78 \\
\hline
\end{tabular}
\caption{Soccer}
\end{subtable}
\vfill
\begin{subtable}{1.\linewidth}
\small
\centering

\begin{tabular}{c|ccccc}
\hline
\texttt{UID}  & 34 & 40 & 244  & 239  & 240  \\ 
Duration (\texttt{min}) & 51.66 & 45.33 & 44.84 &  43.52 & 42.98\\
\hline
\texttt{UID} & 245 & 243 & 46 & 226 & 24\\
Duration (\texttt{min})  & 42.34 & 42.00  & 37.25 & 33.82 & 27.53 \\
\hline
\texttt{UID}   & 254 & 227 & 256 \\
Duration (\texttt{min})  & 29.90 & 20.04 & 14.86 \\
\hline
\end{tabular}
\caption{Dancing}
\end{subtable}
\caption{Total video durations for evaluated participants in Ego-Exo4D.}\label{tab:supp_stat_egoexo}
\end{table}

We select participants from the original training and validation splits of the Ego-Exo4D dataset and evaluate on their exocentric frames that have available GT 3D annotations. 
Note that we do not use the original test split due to the inaccessibility of ground truth annotations, which prevents us from following standard evaluation protocols in human pose estimation to report root-aligned MPJPE and MPJPE-PA.

To construct our evaluation set, we first select participants from the original validation split whose recordings exceed 10 minutes. 
We then expand this set by including participants from the original training split with the longest recording durations.
During this process, we manually check to remove any participants whose faces are anonymized by squares and verify that all test videos sharing the same \texttt{UID} correspond to the same person. 
Furthermore, we skip frames suspected of low annotation quality: 
1) frames with fewer than 10 valid joint annotations; 
2) frames missing left or right hip annotations, as we define the root joint as their midpoint to compute MPJPE; 
3) frames where the average distance between projected 3D annotations and 2D annotations exceeds 20 pixels.

As a result, our evaluation set consists of 13 basketball participants, 4 soccer participants, and 13 dancing participants, totaling 30 participants. 
The imbalance in soccer scenarios is due to the limited number of soccer participants available in the original dataset. 
We report the \texttt{UID} and total durations for selected participants in \cref{tab:supp_stat_egoexo}.

\section{Comparison with Related Works}\label{sec:supp_comparison}

\subsection{Comparison with Domain Generalization Methods}

\begin{table*}[!t]
\small
\centering
\resizebox{0.99\linewidth}{!}{
\begin{tabular}{c|c|c|cc|cc|cc|cc}
\midrule
& \multirow{2}{*}{\makecell{Image\\Encoder\\Backbone}} & \multirow{2}{*}{\makecell{Pre-trained\\ only on \\ Human3.6M}}   &   \multicolumn{2}{c|}{Basketball} &  \multicolumn{2}{c|}{Soccer} &  \multicolumn{2}{c|}{Dance}  &  \multicolumn{2}{c}{All Scenarios} \\
\cline{4-11}
& &&    MPJPE & \makecell{MPJPE\\-PA} &  MPJPE & \makecell{MPJPE\\-PA}&  MPJPE &\makecell{MPJPE\\-PA} &  MPJPE &\makecell{MPJPE\\-PA} \\
\midrule
\multicolumn{11}{c}{\textbf{Domain Generalization Methods (Image-Based)}} \\
\midrule
SPIN~\cite{kolotouros2019learning} & Resnet-50 & &  141.0 & 75.8 & 126.7 & 65.2 & 136.8 & 69.2 & 137.3 & 71.5 \\
I2L-Mesh~\cite{moon2020i2l} & Resnet-50 & & 148.2 & 79.3 & 121.1 & 68.6 & 129.7 & 70.0 & 136.6 & 73.8 \\
PyMAF~\cite{zhang2021pymaf} & Resnet-50 & & 143.0 & 74.5 & 122.5 & 62.7 & 132.0 & 68.2 & 135.5 & 70.2 \\
HMR-2.0b~\cite{goel2023humans} &  ViT-H/16 & & 132.9 & 68.5 & 107.1 & 56.2 & 123.0 & 65.5 & 125.2 & 65.6 \\
TokenHMR~\cite{dwivedi2024tokenhmr} & ViT-H/16 & & 118.2 & 61.7 & 98.6 & 51.1 & 101.5 & 54.4 & 108.4 & 57.1 \\
\midrule
\multicolumn{11}{c}{\textbf{Domain Generalization Methods (Video-Based)}} \\
\midrule
VIBE~\cite{kocabas2020vibe} & Resnet-50 & & 138.6 & 74.2 & 119.3 & 65.3 & 126.1 & 67.5 & 130.6 & 70.1 \\
TCMR~\cite{choi2021beyond} & Resnet-50  & & 135.4 & 71.0 & 114.1 & 62.6 & 121.2 & 64.2 & 126.4 & 66.9 \\
MPS-Net~\cite{wei2022capturing}$^\ddag$ & Resnet-50 & & 132.2 & 70.5 & 108.8 & 60.7 & 111.6 & 63.4 & 120.2 & 66.1 \\
GLoT~\cite{shen2023global} & Resnet-50 & & 135.3 & 70.5 & 114.6 & 62.5 & 122.8 & 64.1 & 127.1 & 66.7 \\
PMCE~\cite{you2023co}$^\ddag$ & Resnet-50 & &  164.4 & 79.5 & 119.2 & 64.8 & 112.0 & 66.0 & 135.7 & 71.7 \\
\midrule
\multicolumn{11}{c}{\textbf{Online Test-Time Adaptation Methods (w/ OpenPose)}} \\
\midrule
BOA~\cite{guan2021bilevel} &Resnet-50 &$\checkmark$  & 158.4 & 72.8 & 115.3 & 63.3 & 117.8 & 69.3 & 135.1 & 70.0 \\
DynaBOA~\cite{guan2022out} &Resnet-50 &$\checkmark$ & 173.0 & 73.4 & 129.1 & 68.2 & 141.0 & 70.9 & 153.3 & 71.6 \\
CycleAdapt~\cite{nam2023cyclic} & Resnet-50 &$\checkmark$ & 160.4 & 80.8 & 126.3 & 78.5 & 135.3 & 81.0 & 145.0 & 80.5 \\
Ours &Resnet-50 &$\checkmark$ & \textbf{141.6} & \textbf{71.3} & \textbf{105.3} & \textbf{61.3}  & \textbf{106.4} & \textbf{67.0} & \textbf{121.5} & \textbf{68.1} \\
\midrule
\multicolumn{11}{c}{\textbf{Online Test-Time Adaptation Methods (w/ ViTPose)}} \\
\midrule
BOA~\cite{guan2021bilevel} &Resnet-50 &$\checkmark$ & 149.6 & 65.3 & 108.9 & 58.2 & 101.9 & 61.9 & 123.5 & 62.9 \\
DynaBOA~\cite{guan2022out} &Resnet-50 &$\checkmark$ & 154.8 & 64.3 & 114.8 & 58.6 & 108.5 & 61.3 & 129.4 & 62.2 \\
CycleAdapt~\cite{nam2023cyclic} &Resnet-50 &$\checkmark$  & 144.8 & 69.6 & 112.1 & 62.9 & 108.3 & 67.1 & 124.6 & 67.6 \\
Ours &Resnet-50 &$\checkmark$ & \textbf{137.4} & \textbf{63.4} & \textbf{102.2} & \textbf{56.8} & \textbf{99.8} & \textbf{59.5} & \textbf{116.4} & \textbf{60.8} \\
\midrule
\end{tabular}}
\caption{Comparison on Ego-Exo4D dataset. All compared methods do not leverage Ego-Exo4D as a training set. $\ddag$ indicates using 3DPW for pre-training.}\label{tab:egoexo}
\end{table*}

\begin{table*}[!t]
\vspace{1.cm}
\small
\centering
\resizebox{0.8\linewidth}{!}{
\begin{tabular}{c|c|c|ccc}
\midrule
& \makecell{Image Encoder\\Backbone} & \makecell{Pre-trained only \\ on Human3.6M} &  MPJPE & MPJPE-PA & MPVPE \\
\midrule
\multicolumn{6}{c}{\textbf{Domain Generalization Methods (Image-Based)}} \\
\midrule
HMR~\cite{kanazawa2018end} & Resnet-50 & & 130.0 & 76.7 & - \\
SPIN~\cite{kolotouros2019learning} & Resnet-50 & & 96.9 & 59.2 & 116.4 \\
I2L-MeshNet~\cite{moon2020i2l}  & Resnet-50 & & 93.2 & 57.7 &110.1\\
PyMAF~\cite{zhang2021pymaf} & Resnet-50 & & 92.8 & 58.9 & 110.1 \\
HMR-2.0b~\cite{goel2023humans} &  ViT-H/16 & & 81.3 & 54.3 & 94.4 \\
TokenHMR~\cite{dwivedi2024tokenhmr} & ViT-H/16  & & 71.0 & 44.3  & 84.6\\
\midrule
\multicolumn{6}{c}{\textbf{Domain Generalization Methods (Video-Based)}} \\
\midrule
HMMR~\cite{kanazawa2019learning} & Resnet-50 && 116.5 & 72.6 & 139.3 \\
VIBE~\cite{kocabas2020vibe} & Resnet-50 && 93.5 & 56.5 & 113.4 \\
TCMR~\cite{choi2021beyond} & Resnet-50 && 95.0 & 55.8 & 111.3 \\
MPS-Net~\cite{wei2022capturing} & Resnet-50 && 91.6 & 54.0 & 109.6 \\
GLoT~\cite{shen2023global} & Resnet-50 && 89.9 & 53.5 & 107.8 \\
PMCE~\cite{you2023co} & Resnet-50 && 81.6 & 52.3 & 99.5 \\
\midrule
\multicolumn{6}{c}{\textbf{Online Test-Time Adaptation Methods (w/ OpenPose)}} \\
\midrule
BOA~\cite{guan2021bilevel} &Resnet-50 &$\checkmark$ &  98.2 & 55.8 & 114.2 \\
DynaBOA~\cite{guan2022out} &Resnet-50 &$\checkmark$& 139.7 & 63.8 & 155.1\\
CycleAdapt~\cite{nam2023cyclic} &Resnet-50 &$\checkmark$& 141.0 & 79.6 & 155.6  \\
Ours &Resnet-50 &$\checkmark$&  \textbf{83.9} & \textbf{51.6} & \textbf{100.3} \\
\midrule
\multicolumn{6}{c}{\textbf{Online Test-Time Adaptation Methods (w/ ViTPose)}} \\
\midrule
BOA~\cite{guan2021bilevel} &Resnet-50 &$\checkmark$& 91.5 & 53.9 & 105.5\\
DynaBOA~\cite{guan2022out} &Resnet-50 &$\checkmark$& 118.4 & 56.1 & 134.5\\
CycleAdapt~\cite{nam2023cyclic} &Resnet-50&$\checkmark$ & 109.6 & 66.6 & 124.8\\
Ours &Resnet-50 &$\checkmark$ & \textbf{85.0} & \textbf{53.3} & \textbf{100.4} \\
\midrule
\end{tabular}}
\caption{Comparison on 3DPW dataset. All compared methods do not leverage 3DPW as a training set.}\label{tab:3dpw}
\end{table*}

\paragraph{Baselines} 
To provide a more comprehensive comparison within the broader scope of 3D human pose estimation, we include additional learning-based works that address the problem from the perspective of domain generalization. 
These methods directly apply the pre-trained model to the test domain without further updating the model. 
Nonetheless, compared to ours, these methods pre-train their models on extensive datasets beyond Human3.6M, where recent works~\cite{dwivedi2024tokenhmr,goel2023humans} further enhance by building on more powerful ViT backbones.

Specifically, we evaluate baseline methods on Ego-Exo4D by leveraging their public codes and released pre-trained weights, and report the results in \cref{tab:egoexo}.
For evaluation on 3DPW, we refer to the number reported in the original paper of baseline methods, and summarize the results in \cref{tab:3dpw}.
Note that for each of these two datasets, none of the compared methods utilizes the corresponding test dataset as pre-training data. 
Moreover, when evaluating baseline methods on the Ego-Exo4D dataset, we use their public pre-trained weights that exclude 3DPW, where applicable. 
However, since MPS-Net~\cite{wei2022capturing} and PMCE~\cite{you2023co} do not release such weights, we use their weights that include 3DPW in pre-training.

\begin{figure*}[!t]
\centering
\includegraphics[width=.99\linewidth]{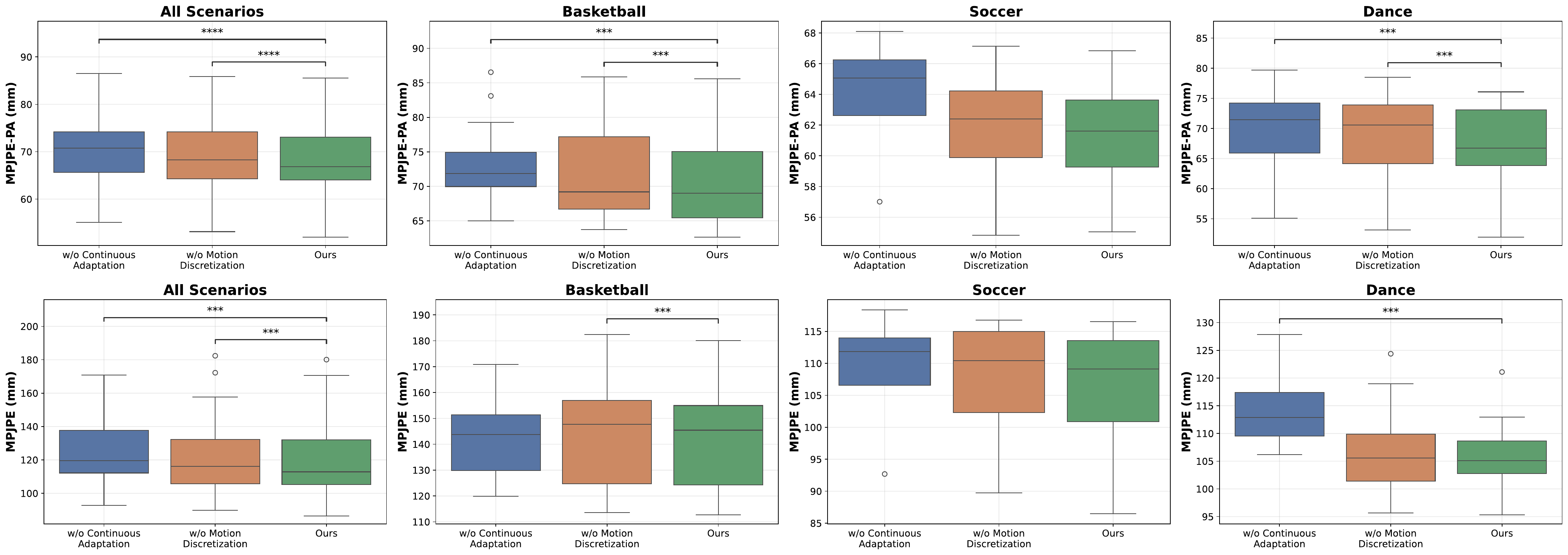}
\caption{Distributional analysis and Wilcoxon signed-ranked tests on 30 evaluated participants from the Ego-Exo4D dataset.
The comparison includes the baseline \textit{w/o continuous adaptation} and \textit{w/o discretization} that are discussed in \cref{fig:error_vs_time} of the main text. 
Statistical significance is indicated as: * $p<0.05$, ** $p<0.01$, *** $p<0.001$, **** $p<0.0001$}\label{fig:supp_wilcoxon}
\end{figure*}

\paragraph{Results}
We first compare our solution with those that also utilize a ResNet-50 backbone. 
As shown in \cref{tab:egoexo,tab:3dpw}, ours demonstrates competitive accuracy. 
On the Ego-Exo4D dataset, our method using ViTPose outperforms all reported baselines on average.
Meanwhile, our method using OpenPose outperforms most reported baselines, particularly in terms of MPJPE.
We empirically observe that the domain gap introduced by undistorted fisheye recordings presents challenges for baseline methods to generalize well, where our adaptation effectively captures the test-time traits to mitigate this domain gap.
On 3DPW, by using 2D detections obtained from OpenPose, we achieve results comparable to PMCE~\cite{you2023co} while outperforming the rest.

We then examine the comparison with HMR-2.0b~\cite{goel2023humans} and TokenHMR~\cite{dwivedi2024tokenhmr}. 
Unlike our solution, these two works leverage more extensive training data and build their backbones on a more powerful modern architecture of ViT. 
Surprisingly, we find our performance on Ego-Exo4D to be competitive with HMR-2.0b, and we even outperform it on average when using 2D detections obtained from ViTPose. 
We also outperform HMR-2.0b on 3DPW regarding the metric of MPJPE-PA and report competitive results when comparing MPJPE. 
Nonetheless, we acknowledge the superior performance of TokenHMR, which we attribute to its exploitation of more extensive training data and more careful designs for motion representation and training loss compared to HMR-2.0b.

In conclusion, by comparing with existing online test-time adaptation methods and domain generalization methods, we demonstrate the effectiveness of our solution in bridging the gap between training and testing domains. 
While we use a ResNet-50 backbone to illustrate the challenge of error accumulation and facilitate fair comparison with existing test-time adaptation methods, we consider integrating modern ViT-based 3D pose estimators as a potential and practical extension to our solution, which we leave as important future work.

\subsection{Qualitative Comparison with Online Adaptation Methods}

We further qualitatively compare our method with BOA~\cite{guan2021bilevel}, DynaBOA~\cite{guan2022out}, and CycleAdapt~\cite{nam2023cyclic} in the supplementary video.
Our approach enables more natural and accurate motion outputs, even under challenges such as occlusion and varying environmental conditions. 
In contrast, both CycleAdapt and DynaBOA suffer from error accumulation as adaptation progresses, failing to effectively leverage test-time personal traits for robust estimation. 
Additionally, BOA and DynaBOA outputs unnatural motion with increased jitter and inaccurate depth estimation (\ie, along the $z$-axis) due to a lack of reliable 3D guidance during adaptation, as demonstrated in the cases of 3DPW and basketball scenario.
These qualitative results are consistent with the quantitative comparisons presented in our main text.

\subsection{Run-time Analysis}

\begin{table}[!t]
\small
\centering
\resizebox{0.99\linewidth}{!}{
\begin{tabular}{c|cccc}
\hline
&  \makecell{BOA\\~\cite{guan2021bilevel}} & \makecell{DynaBOA\\~\cite{guan2022out}} & \makecell{CycleAdapt\\~\cite{nam2023cyclic}} & Ours\\
\hline
\makecell{Time \\ (unit: \SI{}{\second})} & 25.6 & 61.8 & 6.0 & 6.8 \\
\hline
\end{tabular}}
\caption{Running time cost for adapting to each image batches $\mathcal{V}$. 
$\mathcal{V}$ is approximately \SI{5.3}{\second}, containing 160 frames at 30 fps.}\label{tab:runtime}
\end{table}
We test all methods on the same machine, which is equipped with an Intel i9-12900K 5.2 GHz CPU and an Nvidia RTX 3080Ti GPU with 12GB of memory. 
We report the running time costs in \cref{tab:runtime}, where we exclude the time required for 2D detections.

Our method demonstrates competitive efficiency, where we process an image batch $\mathcal{V}$ spanning \SI{5.3}{\second} (\ie, 160 frames at 30 fps) with a cost of \SI{6.8}{\second}.
In contrast, BOA and DynaBOA require significantly more time due to their bilevel optimization process. 
This process involves retrieving pre-trained images for exemplar guidance and sequentially handling each frame by relying on the optimization results of the previous frame.

\section{More Discussion on EgoExo4D}\label{sec:supp_ablation}

\paragraph{Distributional Analysis and Statistical Test}
As shown in \cref{fig:supp_wilcoxon}, we further examine the effects of motion discretization and continuous personalization by conducting statistical analyses on 30 participants from the Ego-Exo4D dataset. 
Specifically, we plot the error distributions for the two ablation baselines and our full solutions, and conduct a Wilcoxon signed-rank test to evaluate statistical significance.

Overall, our method shows statistically significant improvement compared to the baselines across all participants ($p<0.001$). 
While the number of participants per scenario is limited, which may reduce the statistical power and result in non-significant $p$-values in some cases, our method generally shows lower errors, as illustrated in the box plots.

\subsection{Discussion on Basketball Scenarios}

\begin{figure}[!t]
\centering
\includegraphics[width=1.\linewidth]{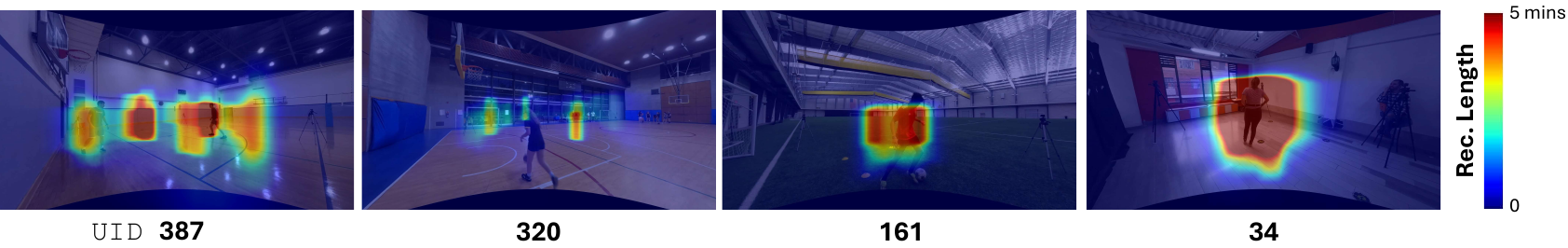}
\caption{Visualization of 2D GT distributions for different participants.
For each pixel, we visualize its frequency of occurrence within the bounding box defined by the 2D GT body joints.}\label{fig:supp_2D_distribution}
\end{figure}

\begin{figure}[!t]
\centering
\includegraphics[width=.99\linewidth]{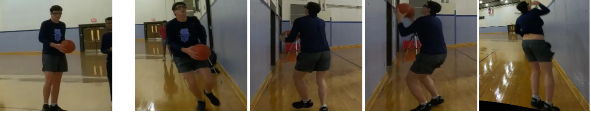}
\caption{Image crops for participant \texttt{UID} 387. 
We show examples with centered (leftmost) and off-center (right) locations in undistorted fisheye camera images. 
Off-center positioning introduces appearance distortion, presenting challenges for adaptation.}\label{fig:supp_387_distort}
\end{figure}

As illustrated in \cref{sec:ablation,fig:error_vs_time} of the main text, although we demonstrate benefits with continuous personalized adaptation on average, we observe limited enhancement in basketball scenarios.
Furthermore, as reported in \cref{tab:supp_per_sbj}, while our continuous adaptation yields clear benefits for some basketball participants (\eg, 323, 318, 320, 316, 382, 317), it results in larger errors for others (\eg, 386, 387, 383, 376), especially in terms of MPJPE.

We attribute this phenomenon to the following reasons. 
As shown in \cref{fig:supp_2D_distribution}, participants who do not benefit from continuous adaptation often appear off-center in recordings, as exemplified by the participant with \texttt{UID} 387. 
This off-center positioning causes significant appearance distortion when undistorting fisheye recordings (see \cref{fig:supp_387_distort}), which further enlarges the train-test domain gap.
This increased domain gap complicates adaptation, resulting in larger batch-wise errors even when starting from pre-trained weights for each incoming image batch $\mathcal{V}$.
Consequently, mitigating this domain gap and correcting accumulated errors becomes highly challenging, which we recognize as a critical aspect for future work.

\begin{table}[t]
\small
\centering
\resizebox{0.99\linewidth}{!}{
\begin{tabular}{c|cc|cc|cc}
\hline
\multirow{2}{*}{\texttt{UID}} & \multicolumn{2}{c|}{\makecell{w/o Continuous\\Adaptation}} & \multicolumn{2}{c|}{\makecell{w/o Motion \\Discretization}} & \multicolumn{2}{c}{Ours-Full} \\
\cline{2-7}
& MPJPE & \makecell{MPJPE-PA} & MPJPE & \makecell{MPJPE-PA} & MPJPE & \makecell{MPJPE-PA} \\
\hline
 388 & \textbf{144.8} & 70.9 & 149.3 & 67.3 & 146.4 & \textbf{66.8} \\
 386 & \textbf{143.8} & \textbf{72.7} & 151.5 & 74.3 & 149.0 & 73.0 \\
 387 & \textbf{159.6} & 70.7 & 172.2 & 69.0 & 170.6 & \textbf{67.9} \\
 384 & \textbf{145.4} & \textbf{74.9} & 147.7 & 77.2 & \textbf{145.4} & 75.0 \\
 383 & \textbf{170.8} & 75.0 & 182.4 & 73.9 & 180.1 & \textbf{73.0} \\
 323 & 119.9 & 65.8 & 113.7 & 64.8 & \textbf{112.7} & \textbf{64.6} \\
 318 & 138.4 & 83.1 & 132.6 & 83.0 & \textbf{132.3} & \textbf{81.9} \\
 376 & \textbf{151.4} & 86.5 & 156.9 & 85.9 & 155.0 & \textbf{85.6} \\
 320 & 129.8 & 71.9 & 124.8 & 69.2 & \textbf{124.2} & \textbf{69.0} \\
 316 & 125.8 & 70.0 & 120.5 & 65.7 & \textbf{118.6} & \textbf{65.4} \\
 382 & 122.2 & 65.0 & 118.7 & 63.8 & \textbf{118.3} & \textbf{62.6} \\
 558 & \textbf{156.2} & 65.6 & 157.6 & 66.7 & 156.5 & \textbf{64.8} \\
 317 & 135.6 & 79.3 & 131.4 & \textbf{77.3} & \textbf{131.3} & \textbf{77.3} \\
\hline
 Mean & 141.8 & 73.2 & 143.0 & 72.2 & \textbf{141.6} & \textbf{71.3} \\
\hline
\end{tabular}}
\caption{Results on each basketball participants of Ego-Exo4D dataset, with 2D detections provided by OpenPose. We compare our complete solution with two baselines discussed in \cref{fig:error_vs_time} of the main text.}\label{tab:supp_per_sbj}
\end{table}

Nonetheless, \cref{tab:supp_per_sbj} verifies that our continuous personalized adaptation benefits basketball participants who more frequently appear around the image center (\eg, \cref{fig:supp_2D_distribution}-\texttt{UID} 320).
Meanwhile, \cref{fig:supp_wilcoxon} demonstrates the benefits of continuous adaptation for participants in soccer and dancing scenarios. 
Additionally, our proposed motion discretizations show clear advantages in mitigating error accumulation for enhanced performance.

\begin{table*}[!t]
\small
\centering
\vfill
\begin{subtable}{1.0\textwidth}
\centering
\resizebox{1.0\textwidth}{!}{
\begin{tabular}{c|cc|cc|cc|cc|cc}
\hline
\multirow{2}{*}{\makecell{w/ Soft-Reset}} &  \multicolumn{2}{c|}{w/ Motion Discretization} & \multicolumn{2}{c|}{Basketball} &  \multicolumn{2}{c|}{Soccer} &  \multicolumn{2}{c|}{Dance}  &  \multicolumn{2}{c}{All Scenarios} \\
\cline{2-3}
& \makecell{w/ $L_{ach}$ ($\theta^\ast$)} & \makecell{w/ Self-Replay} & MPJPE & MPJPE-PA &  MPJPE & MPJPE-PA &  MPJPE & MPJPE-PA &  MPJPE & MPJPE-PA \\
\hline
&&& 159.4 & 81.2 & 125.3 & 77.9 & 135.0 & 80.8 & 144.3 & 80.6 \\
$\checkmark$ & & &   143.0 & 72.2 & 106.8 & 61.7 & 107.6 & 68.6 & 122.9 & 69.2 \\
& $\checkmark$  & $\checkmark$  &  154.5 & 76.2 & 118.4 & 70.2 & 127.9 & 74.7 & 138.2 & 74.8 \\
$\checkmark$  & $\checkmark$ & $\checkmark$ & \textbf{141.6} & \textbf{71.3} & \textbf{105.3} & \textbf{61.3} & \textbf{106.4} & \textbf{67.0} & \textbf{121.5} & \textbf{68.1} \\
\hline
$\checkmark$ & & & 143.0 & 72.2 & 106.8 & 61.7 & 107.6 & 68.6 & 122.9 & 69.2 \\
$\checkmark$ & $\checkmark$ & &144.6 & 72.0 & 105.9 & 61.7 & 107.2 & 68.7 & 123.2 & 69.2 \\
$\checkmark$ & & $\checkmark$ & 143.0 & 72.1 & 106.9 & 61.6 & 107.7 & 68.5 & 122.9 & 69.2 \\
$\checkmark$ & $\checkmark $ & $\checkmark$ &  \textbf{141.6} & \textbf{71.3} & \textbf{105.3} & \textbf{61.3} & \textbf{106.4} & \textbf{67.0} & \textbf{121.5} & \textbf{68.1} \\
\hline
\end{tabular}}
\caption{Discussion on soft reset and motion discretization}
\end{subtable}

\vfill
\begin{subtable}{1.0\textwidth}
\small
\centering
\begin{tabular}{cc|cc|cc|cc|cc}
\hline
\multirow{2}{*}{\makecell{w/ $L_p$ \\($\theta^\prime$)}} &\multirow{2}{*}{\makecell{w/ $L_{ach}$\\ ($\theta^\ast)$}} &  \multicolumn{2}{c|}{Basketball} &  \multicolumn{2}{c|}{Soccer} &  \multicolumn{2}{c|}{Dance}  &  \multicolumn{2}{c}{All Scenarios} \\
&  & MPJPE & MPJPE-PA &  MPJPE & MPJPE-PA &  MPJPE & MPJPE-PA &  MPJPE & MPJPE-PA \\
\hline
$\checkmark$ & & 143.0 & 72.1 & 106.9 & 61.6 & 107.7 & 68.5 & 122.9 & 69.2 \\
& $\checkmark$  & 143.2 & 73.2 & 106.7 & 62.6 & 108.2 & 68.6 & 123.1 & 69.8 \\
$\checkmark$ & $\checkmark $ &  \textbf{141.6} & \textbf{71.3} & \textbf{105.3} & \textbf{61.3} & \textbf{106.4} & \textbf{67.0} & \textbf{121.5} & \textbf{68.1} \\
\hline
\end{tabular}
\caption{Discussion on 3D supervision signals used in adapting the pose estimator $\mathrm{F}$}\label{tab:supp_discretization_pose}
\end{subtable}

\vfill
\begin{subtable}{1.0\textwidth}
\small
\centering
\begin{tabular}{c|cc|cc|cc|cc}
\hline
\multirow{2}{*}{\makecell{Soft-Reset Decay\\  $\mu_F$}} &  \multicolumn{2}{c|}{Basketball} &  \multicolumn{2}{c|}{Soccer} &  \multicolumn{2}{c|}{Dance}  &  \multicolumn{2}{c}{All Scenarios} \\
&  MPJPE & MPJPE-PA &  MPJPE & MPJPE-PA &  MPJPE & MPJPE-PA &  MPJPE & MPJPE-PA \\
\hline
0  & 154.5 & 76.2 & 118.4 & 70.2 & 127.9 & 74.7 & 138.2 & 74.8 \\
0.9 & 143.6 & 72.1 & 105.7 & 61.9 & 106.9 & 68.5 & 122.7 & 69.2 \\
0.95& \textbf{141.6} & \textbf{71.3} & \textbf{105.3} & \textbf{61.3} & \textbf{106.4} & \textbf{67.0} & \textbf{121.5} & \textbf{68.1} \\
1.0 & 141.8 & 73.2 & 108.7 & 63.8 & 113.8 & 69.1 & 125.3 & 70.2 \\
\hline
\end{tabular}
\caption{Discussion on soft reset decay $\mu_F$ for the pose estimator $\mathrm{F}$. }\label{tab:supp_soft_reset}
\end{subtable}

\vfill
\begin{subtable}{1.0\textwidth}
\small
\centering
\begin{tabular}{c|cc|cc|cc|cc}
\hline
\multirow{2}{*}{\makecell{w/ Continuous \\ Adaptation}}&  \multicolumn{2}{c|}{Basketball} &  \multicolumn{2}{c|}{Soccer} &  \multicolumn{2}{c|}{Dance}  &  \multicolumn{2}{c}{All Scenarios}\\
&  MPJPE & MPJPE-PA &  MPJPE & MPJPE-PA &  MPJPE & MPJPE-PA &  MPJPE & MPJPE-PA \\
\hline
 & 141.8 & 73.2 & 108.7 & 63.8 & 113.8 & 69.1 & 125.3 & 70.2 \\
$\checkmark$ & \textbf{141.6} & \textbf{71.3} & \textbf{105.3} & \textbf{61.3} & \textbf{106.4} & \textbf{67.0} & \textbf{121.5} & \textbf{68.1} \\
\hline
\end{tabular}
\caption{Discussion on continuous adaptation.}\label{tab:supp_reset}
\end{subtable}

\caption{Per-scenario results on Ego-Exo4D for ablation studies and discussions in the main text.}\label{tab:supp_main}
\end{table*}

\begin{table*}[!t]
\small
\centering
\begin{tabular}{c|cc|cc|cc|cc}
\hline
\multirow{2}{*}{ \makecell{Num Layers of $\mathcal{C}$ \\$k$}} &  \multicolumn{2}{c|}{Basketball} &  \multicolumn{2}{c|}{Soccer} &  \multicolumn{2}{c|}{Dance}  &  \multicolumn{2}{c}{All Scenarios} \\
&  MPJPE & MPJPE-PA &  MPJPE & MPJPE-PA &  MPJPE & MPJPE-PA &  MPJPE & MPJPE-PA \\
\hline
0 & 143.0 & 72.2 & 106.8 & 61.7 & 107.6 & 68.6 & 122.9 & 69.2 \\
1 & 141.9 & 71.7 & 105.8 & \textbf{61.3} & 107.0 & 67.4 & 122.0 & 68.5 \\
2 & 141.7 & \textbf{71.3} & 105.4 & 61.4 & 106.7 & 67.1 & 121.7 & 68.2 \\
3 & \textbf{141.6} & \textbf{71.3} & \textbf{105.3} & \textbf{61.3} & \textbf{106.4} & \textbf{67.0} & \textbf{121.5} & \textbf{68.1} \\
\hline
\end{tabular}
\caption{Discussion on the number of codebook layers. We report results on Ego-Exo4D.}\label{tab:supp_ncb}
\end{table*}
\begin{table*}[!t]
\small
\centering
\begin{tabular}{c|cc|cc|cc|cc}
\hline
\multirow{2}{*}{\makecell{w/ Updating $\mathcal{C}$}} &  \multicolumn{2}{c|}{Basketball} &  \multicolumn{2}{c|}{Soccer} &  \multicolumn{2}{c|}{Dance}  &  \multicolumn{2}{c}{All Scenarios} \\
&  MPJPE & MPJPE-PA &  MPJPE & MPJPE-PA &  MPJPE & MPJPE-PA &  MPJPE & MPJPE-PA \\
\hline
& 142.1 & \textbf{71.2} & \textbf{105.3} & \textbf{61.3} & \textbf{106.3} & 67.3 & 121.7 & 68.2 \\
$\checkmark$& \textbf{141.6} & 71.3 & \textbf{105.3} & \textbf{61.3} & 106.4 & \textbf{67.0} & \textbf{121.5} & \textbf{68.1} \\
\hline
\end{tabular}
\caption{Ablation study on updating codebook $\mathcal{C}$ using the replay latent $\bar{\vb*{z}}$ during adaptation. We report results on Ego-Exo4D.}\label{tab:supp_update_codebook}
\end{table*}

\begin{table*}[!t]
\small
\centering
\begin{tabular}{ccc|cc|cc|cc|cc}
\hline
\multirow{2}{*}{\makecell{\\ w/ Motion\\Discretization}} &  \multirow{2}{*}{\makecell{\\ w/ Soft Reset \\ on $\mathrm{M}$}} & \multirow{2}{*}{\makecell{\\ Soft-Reset \\ Decay $\mu_M$}} & \multicolumn{2}{c|}{Basketball} &  \multicolumn{2}{c|}{Soccer} &  \multicolumn{2}{c|}{Dance}  &  \multicolumn{2}{c}{All Scenarios} \\
&&&   MPJPE & \makecell{MPJPE\\-PA} &  MPJPE &  \makecell{MPJPE\\-PA} &  MPJPE &  \makecell{MPJPE\\-PA} &  MPJPE & \makecell{MPJPE\\-PA} \\
\hline
&&-& 143.0 & 72.2 & 106.8 & 61.7 & 107.6 & 68.6 & 122.9 & 69.2 \\
&$\checkmark$ & 0.95 & 142.7 & 72.0 & 107.0 & 61.6 & 107.7 & 68.4 & 122.8 & 69.1 \\
&$\checkmark$ & 1.0 & 142.5 & 72.1 & 106.6 & 61.6 & 107.2 & 68.0 & 122.4 & 68.9 \\
$\checkmark$ & & - & \textbf{141.6} & \textbf{71.3} & \textbf{105.3} & \textbf{61.3} & \textbf{106.4} & \textbf{67.0} & \textbf{121.5} & \textbf{68.1} \\
\hline
\end{tabular}
\caption{Comparison of motion discretization with an alternative solution that also applies the soft reset to $\mathrm{M}$. 
We report results on Ego-Exo4D.}\label{tab:supp_soft_reset_m}
\end{table*}

\subsection{Discussion on Ablation Study}

\paragraph{Per-Scenario Results for Main Text Tables}
In \cref{tab:supp_main}, we present the per-scenario results for the ablation studies and discussions conducted in our main text.
These results further support the discussion in the main text and demonstrate the performance gain of our key designs across scenarios.

\paragraph{Residual codebook benefits.}
In \cref{tab:supp_ncb}, we further examine the residual design of our codebook $\mathcal{C}$ by discussing the number of layers $k$.
Specifically, $k=0$ indicates that $\mathcal{C}$ is not used, while $k=1$ introduces motion discretization but without a residual design.

We observe that the residual design enhances effectiveness, particularly in dance scenarios that involve a more diverse range of motion. 
The residual design efficiently captures motion patterns at both coarse and detailed levels, enabling the representation of $(N_c)^k$ patterns with a memory cost of $kN_c$ codes.

\paragraph{Codebook Update in Adaptation}
As described in \cref{sec:replay} and \cref{alg:adaptation}, when adapting the motion denoising network $\mathrm{M}$, we update the codebook $\mathcal{C}$ with the latent $\vb*{\bar{z}}$, with $\vb*{\bar{z}}$ encoding the replay motions $\vb*{\bar{\theta}}$.
We ablate this design and report the results in \cref{tab:supp_update_codebook}. 
We observe a moderate average performance gain with this design, especially in terms of the MPJPE in basketball scenarios.

\paragraph{Comparison with Soft Reset on $\mathrm{M}$}
Our proposed solution utilizes motion discretization to mitigate the effects of continuously adapting the motion denoising network $\mathrm{M}$ with imperfect estimations, while applying the soft reset only to the pose estimator $\mathrm{F}$.
Alternatively, we notice that one may mitigate this impact using simpler approaches:
for instance, by applying the soft reset to $\mathrm{M}$ as well, or by resetting $\mathrm{M}$ to its pre-trained weights after adaption on each batch (\ie, with decay $\mu_M=1$ in the soft reset).

We compare and report these solutions in \cref{tab:supp_soft_reset_m}.
The results verify our design choice, demonstrating that our motion discretization (last row of \cref{tab:supp_soft_reset_m}) effectively exploits historical adaptation while mitigating the challenge of continuous self-supervised adaptation with imperfect estimations.


\end{document}